\documentclass[sigconf,nonacm]{acmart}
\usepackage{subfiles}
\usepackage{booktabs}
\usepackage{multirow}
\usepackage{stfloats}
\usepackage{lipsum}
\usepackage{arydshln}
\usepackage{graphicx}
\usepackage{subcaption}
\usepackage[normalem]{ulem}

\usepackage{amssymb}
\usepackage{tikz}
\usepackage{amsmath}
\usepackage{enumitem}
\usepackage{xcolor}

\definecolor{mycolor_purple}{RGB}{112, 48, 160}
\definecolor{mycolor_orange}{RGB}{255, 102, 0}
\definecolor{mycolor_green}{RGB}{83, 129, 53}
\definecolor{mycolor_blue}{RGB}{48, 84, 151}

 \newcommand{\vic}[1]{\textcolor{black}{#1}}
\newcommand{\zwadd}[1]{\textcolor{black}{#1}}

\AtBeginDocument{%
  \providecommand\BibTeX{{%
    \normalfont B\kern-0.5em{\scshape i\kern-0.25em b}\kern-0.8em\TeX}}}

\setcopyright{acmlicensed}
\copyrightyear{2025}
\acmYear{2025}
\acmDOI{XXXXXXX.XXXXXXX}

\acmConference[KDD '25]{Make sure to enter the correct
  conference title from your rights confirmation emai}{August 03--07,
  2025}{Toronto, Canada}
\acmBooktitle{Woodstock '18: ACM Symposium on Neural Gaze Detection,
 August 03--07, 2025, Toronto, Canada} 
\acmISBN{978-1-4503-XXXX-X/18/06}

\begin{document}

\title{Multi-level Matching Network for Multimodal Entity Linking}
\author{Zhiwei Hu}
\affiliation{
  \institution{School of Computer and Information Technology \\ Shanxi University}
  \city{Taiyuan}
  \country{China}
}
\email{zhiweihu@whu.edu.cn}

\author{Víctor Gutiérrez-Basulto}
\authornote{Contact Authors.}
\affiliation{
  \institution{School of Computer Science and Informatics\\ Cardiff University}
  \city{Cardiff}
  \country{UK}
}
\email{gutierrezbasultov@cardiff.ac.uk}

\author{Ru Li}
\authornotemark[1]
\affiliation{
  \institution{School of Computer and Information Technology \\ Shanxi University}
  \city{Taiyuan}
  \country{China}
}
\email{liru@sxu.edu.cn}

\author{Jeff Z. Pan}
\authornotemark[1]
\affiliation{
  \institution{ILCC, School of Informatics\\ University of Edinburgh}
  \city{Edinburgh}
  \country{UK}
}
\email{j.z.pan@ed.ac.uk}






\begin{abstract}
\vic{Multimodal entity linking (MEL) aims to link ambiguous mentions within multimodal contexts to corresponding entities in a multimodal knowledge base. Most existing approaches to MEL are based  on  representation learning or vision-and-language pre-training mechanisms for exploring the complementary effect among multiple modalities. However, these methods  suffer from two  limitations. On the one hand, they overlook the possibility of considering negative samples from the same modality. On the other hand, they lack  mechanisms to capture  bidirectional cross-modal interaction. To address these issues, we propose a  \textbf{M}ulti-level \textbf{M}atching network for \textbf{M}ultimodal \textbf{E}ntity \textbf{L}inking (${\rm \textbf{M}^\textbf{3}\textbf{EL}}$). Specifically, ${\rm M^3EL}$ is composed of three different modules: (\textit{i} ) a \textit{Multimodal Feature Extraction} module, which extracts modality-specific representations with  a multimodal encoder and introduces an intra-modal contrastive learning sub-module to obtain  better discriminative embeddings  based on uni-modal differences; (\textit{ii}) an \textit{Intra-modal Matching Network} module, which contains two levels  of matching granularity: \textit{Coarse-grained Global-to-Global}  and \textit{Fine-grained Global-to-Local}, to achieve local and global level intra-modal interaction; (\textit{iii}) a \textit{Cross-modal Matching Network} module, which applies bidirectional strategies,  \textit{Textual-to-Visual} and \textit{Visual-to-Textual} matching, to implement bidirectional cross-modal interaction. Extensive experiments conducted on WikiMEL, RichpediaMEL, and WikiDiverse datasets demonstrate the outstanding performance of ${\rm M^3EL}$ when compared to the state-of-the-art baselines. 
}

\end{abstract}

\maketitle

\section{Introduction}
\begin{figure}[t!]
    \centering
    \includegraphics[width=0.48\textwidth]{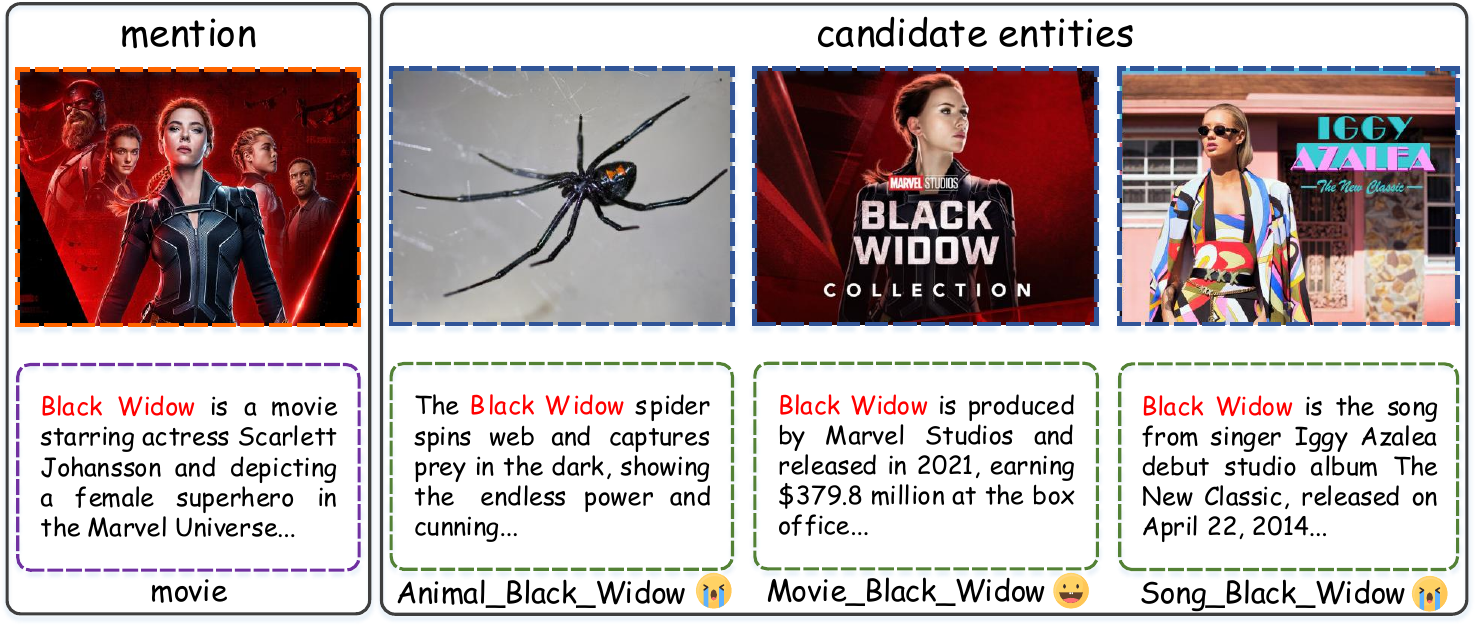}
    \caption{\vic{An example of MEL. Dotted boxes of different colors represent different features: color \textcolor{mycolor_purple}{purple} for mention textual description (mention text), color \textcolor{mycolor_orange}{orange} for mention visual context (mention image), color \textcolor{mycolor_green}{green} for entity textual description (entity text), color \textcolor{mycolor_blue}{blue} for entity visual context (entity image).}}
    \label{figure_instance}
\end{figure}

\vic{Entity Linking (EL) aims at aligning the mentions within a context to the corresponding entities in a knowledge base~\citep{Johannes_2011, Wei_2015}. EL supports numerous downstream information-retrieval applications, such as question answering~\citep{Wen_2015, Wenhan_2019, Shayne_2021, Zhiwei_2022, Yan_2024}, semantic search~\citep{Tao_2007, Ilaria_2013, Bowei_2023}, dialog systems~\citep{Lizi_2018, Ali_2019}, and so on~\citep{Hu_2024, Han_2024}. Most existing EL frameworks  focus on mention disambiguation  via context resolution in  the textual modality. However, for  multimodal information including images along with text,  conventional text-based approaches to  EL struggle to effectively encode such complex  content. Multimodal Entity Linking (MEL) extends traditional EL by considering multimodal information, i.e., it aims at linking textual and visual mentions into their corresponding entities in a multimodal knowledge base. For example, in Figure~\ref{figure_instance}, the mention \textit{Black Widow} can be linked to the  entities \textit{Animal\_Black\_Widow}, \textit{Movie\_Black\_Widow} and \textit{Song\_Black\_Widow}. Although, it is possible to roughly guess the  entity to which the mention has to be linked by relying solely on the textual description, the associated  visual  information  can be used to improve the confidence of text-based  predictions. Indeed,  the image associated  to  the mention \textit{Black Widow} has a high degree of semantic similarity with that of the entity \textit{Movie\_Black\_Widow},  both including the actress  \textit{Scarlett Johansson}, which substantially increases the probability that the mention \textit{Black Widow} is linked to the entity \textit{Movie\_Black\_Widow}. }

\vic{A wide variety of approaches to MEL have been already proposed, including representation learning (RL) frameworks~\citep{Seungwhan_2018, Omar_2020, Qiushuo_2022, Peng_2022, Shangyu_2023, Chengmei_2023}, vision-and-language pre-training (VLP) based methods~\citep{Alec_2021, Wonjae_2021, Junnan_2021, Zi_2022, Pengfei_2023}, 
and a generative-based method~\citep{Senbao_2024}. Despite the substantial progress achieved so far, each  of these approaches  have their own disadvantages. The generative-based method  GEMEL~\citep{Senbao_2024} leverages the in-context learning capability of large language models (LLMs) to create demonstrations, which requires a substantial amount of time and space overhead, and its performance is unsatisfactory. RL-based methods explore complementary effects of different modalities via joint or collaborative representation learning, e.g., using the concatenation operation~\citep{Omar_2020}, additive attention~\citep{Seungwhan_2018} or cross-attention mechanisms~\citep{Peng_2022}. However, the above methods fail to integrate the information from pre-trained visual and textual models into the model representation process. VLP-based methods use a model jointly pre-trained on visual and language tasks as a modal encoder, although it can better understand the association between textual and visual information, they still suffer of two problems:}
\begin{enumerate}[itemsep=0.5ex, leftmargin=5mm]
\item \textbf{Diversity of Negative Samples.} 
\vic{To learn  better textual and visual embeddings, VLP models like CLIP~\citep{Alec_2021} introduce a contrastive loss  bringing  paired image-text representations together, while pushing  unpaired instances away from each other. However, they overlook the possibility of using negative samples from the same modality. Indeed,  VLP-based MEL methods directly apply a VLP model as the textual and visual encoder, disregarding  intra-modal negative samples.  However, for the MEL task, both mentions and candidate entities contain independent  textual and visual information. Implementing entity-mention interaction within the textual or visual modality is a natural way to improve  the model representation capabilities.}

\item \textbf{Bidirectional Cross-modal Interaction.} 
\vic{Existing methods only consider a one-way information flow, from textual to visual or visual to textual, during cross-modal interaction, i.e., they lack bidirectional  cross-modal interaction. In practice, the textual-to-visual flow focuses more on the impact of the textual  knowledge on the visual modality, while visual-to-textual pays attention to applying visual  knowledge to the text modality. Therefore, we advocate that it is necessary to design a mechanism that better captures the bidirectional interaction pattern, to fully harness the knowledge of all available modalities.}
\end{enumerate}
\vic{To address the above two shortcomings, we propose a  \textbf{M}ulti-level \textbf{M}atching network for \textbf{M}ultimodal \textbf{E}ntity \textbf{L}inking (${\rm \textbf{M}^\textbf{3}\textbf{EL}}$).  ${\rm M^3EL}$ includes the following three modules. The \textit{Multimodal Feature Extraction} module utilizes a pre-trained CLIP model to obtain modality-specific representations for each entity and mention. To alleviate shortcoming (1), we introduce an \textit{Intra-modal Contrastive Learning} module to include negative samples within a modality into the discriminative embedding acquisition process. Additionally, we employ an \textit{Intra-modal Matching Network} module which contains two levels of matching granularity, \textit{Coarse-grained Global-to-Global}  and \textit{Fine-grained Global-to-Local} matching, to realize the interaction between local and global features within a modality. Moreover, we design a \textit{Cross-modal Matching Network} module which considers bidirectional matching strategies,  \textit{Textual-to-Visual} and \textit{Visual-to-Textual} matching, to reduce the gap between the distribution over different modalities, addressing  shortcoming (2).}

\begin{itemize}[itemsep=0.5ex, leftmargin=5mm]

\item \vic{We propose the ${\rm M^3EL}$ framework for MEL, which simultaneously considers the diversity of negative samples  and bidirectional cross-modal interaction.}
\item \vic{We introduce an intra-modal contrastive learning mechanism to obtain better discriminative embedding representations that are faithful within a modality.}
\item \vic{We devise intra-modal and cross-modal matching networks to  explore different multimodal interactions, reducing the gap over intra-modality and cross-modality distributions. }
\item \vic{We conduct a series of empirical and ablation experiments on three well-known datasets, showing the strong performance of ${\rm M^3EL}$ when compared to  the  state-of-the-art.} Code and data are available at: https://github.com/zhiweihu1103/MEL-M3EL.
\end{itemize}

\section{Related Work}
\smallskip
\noindent\textbf{$\triangleright$ Entity Linking.} 
\vic{Recent methods for Entity Linking (EL) mainly focus on exploiting ambiguous mentions to the referent unambiguous entities in a given knowledge base, which can be divided into two series: \textit{local-level} methods and \textit{global-level} methods. Local-level methods~\citep{Matthew_2019, Ledell_2020, Nicola_2021} primarily consider mention along with its surrounding words or sentence to capture contextual information. Global-level methods~\citep{Octavian_2017, Phong_2018, Chenwei_2018, Yixin_2018, Zheng_2019, Xiyuan_2019} also take entity or topic coherence into account to calculate the mention and entity semantic consistency. However, these methods do not work well when processing multimodal data, including textual and visual content.}


\smallskip
\noindent\textbf{$\triangleright$ Multimodal Entity Linking.} \vic{ Multimodal entity linking is an extension of  the traditional entity linking task that utilizes additional multimodal information (e.g., visual information) to support  the disambiguation of entities. Mainstream approaches can be classified into two categories: Representation Learning (RL) frameworks and Vision-and-Language Pre-training (VLP) methods. (\textit{i}): \textit{RL-based methods}: DZMNED~\citep{Seungwhan_2018} utilizes a multimodal attention mechanism to fuse textual, visual and character features of mentions and entities. JMEL~\citep{Omar_2020} introduces  fully connected layers  to embed the textual and visual information  into an implicit joint space. VELML~\citep{Qiushuo_2022} designs a deep modal-attention neural network to aggregate different modality features and map visual objects to the entities. GHMFC~\citep{Peng_2022} extracts the hierarchical features of textual and visual co-attention through a multi-modal co-attention mechanism. DRIN~\citep{Shangyu_2023} explicitly encodes four different types of alignments between mentions and entities, and builds graph convolutional network to dynamically select the corresponding alignment relations for different input samples. MMEL~\citep{Chengmei_2023} proposes a joint feature extraction module to learn textual and visual representations, and a pairwise training schema and multi-mention collaborative ranking mechanism to model the potential connections. (\textit{ii}): \textit{VLP-based models}: CLIP~\citep{Alec_2021} trains on large-scale image-caption pairs with contrastive self-supervised objectives to attain textual and visual representations. ViLT~\citep{Wonjae_2021} discards convolutional visual features and adopts a vision transformer to model long-range dependencies over a sequence of fixed-size non-overlapping image patches. ALBEF~\citep{Junnan_2021} introduces a contrastive loss to align the textual and visual representations before fusing them through cross-modal attention to enable more grounded vision and language representation learning. METER~\citep{Zi_2022} systematically investigates how to train a full-transformer vision-language pre-trained model in an end-to-end manner. MIMIC~\citep{Pengfei_2023} utilizes BERT~\citep{Jacob_2019} and CLIP as textual and visual encoder, respectively, and organizes three interaction mechanisms to comprehensively  explore the intra-modal and inter-modal interaction among embeddings of entities and mentions. (\textit{iii}): \textit{generative-based method}: GEMEL~\citep{Senbao_2024} leverages the in-context learning capability of LLMs (e.g., LlaMa-2-7B~\citep{Hugo_2023}) by retrieving multimodal instances as demonstrations, and then applies a constrained decoding strategy to efficiently search the valid entity space. Considering that it requires the use of an LLM with intensive computation and parameters, this poses a great challenge to the efficiency of GEMEL. Furthermore, there is a large gap between its performance and the state-of-the-art. Therefore, generative-based method is not mainstream and is still in an exploration phase. Although existing RL-based or VLP-based methods have made significant progress, they still have two limitations that need to be addressed: On the one hand,  VLP-based MEL methods directly utilize VLP as the textual and visual encoder, however VLP does not consider negative samples within a modality during the pre-training process. Both entities and mentions in the MEL task contain textual and visual  knowledge. Considering negative samples of entities and mentions in the same modality is useful for improving contrastive learning capabilities. On the other hand, existing methods only consider the information flow from textual to visual or from visual to textual when interacting between modalities, and lack  bidirectional  cross-modal interaction.}

\section{Method}
\begin{figure*}[!htp]
    \centering
    \includegraphics[width=1\textwidth]{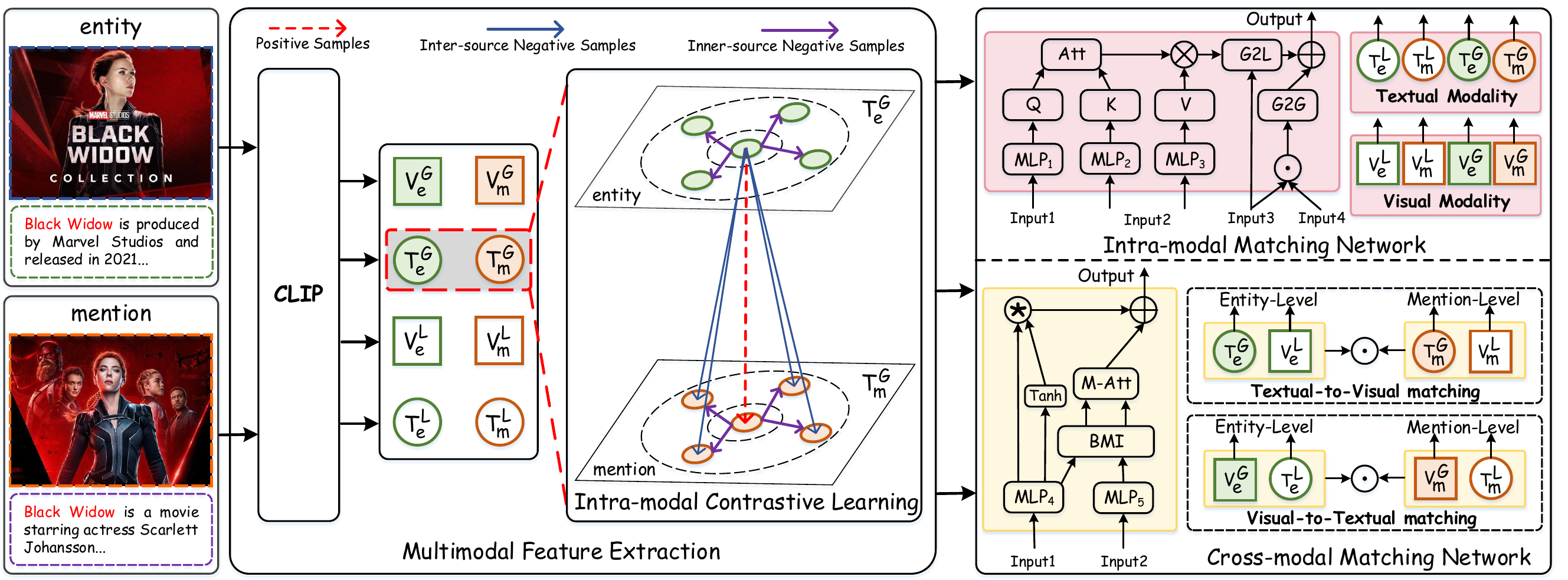}
    \caption{\vic{The  structure of the ${\rm M^3EL}$ model, containing three modules: Multimodal Feature Extraction (MFE) with Intra-modal Contrastive Learning (ICL), Intra-modal Matching Network (IMN) and Cross-modal Matching Network (CMN). \textit{Att} and \textit{M-Att} denote the attention and multi-heads attention mechanisms, respectively.}}
    \label{figure_model}
\end{figure*}

\zwadd{In this section, we first define the task of multimodal entity linking, and then introduce the ${\rm M^3EL}$ framework, including: Multimodal Feature Extraction, Intra-modal Matching Network, and Cross-modal Matching Network.}

\subsection{Task Definition}
\noindent\textbf{$\triangleright$ Multimodal Entity Linking.}
\vic{Let $\mathcal{E}$ be a set of entities in a multimodal knowledge base $\mathcal{K}$. Each entity $E_e \in \mathcal E$ is of the form $\{N_e, T_e, V_e\} $, where $N_e$ denotes the  name of the entity, $T_e$ is a  textual description of the entity and $V_e$ represents the  visual context of the entity associated with its textual description. A \emph{mention}  $E_m$ (and its context)  is of the form $\{N_m, T_m, V_m\}$, where $N_m$, $T_m$, and $V_m$ respectively are the name of  the mention, the token sequence in which the mention is located, and the corresponding visual image of the mention. The \emph{multimodal entity linking (MEL) task} aims to retrieve the ground truth entity $E_e \in \mathcal E$ that is  the most relevant to the mention $E_m$. For example,  in Figure~\ref{figure_instance}, the mention \textit{Black Widow} requires to be  linked to one of the three candidate entities in the set \{\textit{Animal\_Black\_Widow}, \textit{Movie\_Black\_Widow}, \textit{Song\_Black\_Widow}\}. After combining the textual description and visual information, we can  conclude that the entity \textit{Movie\_Black\_Widow} is the most  relevant for the mention. Usually, the MEL task can be formulated by maximizing the log-likelihood over the training set $\mathcal{D}$ as}: 
\begin{equation}
\label{equation_1}
\varrho^*=\underset{\varrho}{{\rm max}}\sum_{(E_m, E_e)\in\mathcal{D}}{\rm log}\,sim(E_e|E_m, \mathcal{E})
\end{equation}

\vic{where $sim(\cdot)$ calculates the similarity between the mention $E_m$ and entity $E_e$, and $\varrho$ represents the parameters involved in the optimization process, $\varrho^*$ denotes the final model.}
\subsection{MFE: Multimodal Feature Extraction} 
\subsubsection{Multi-modal Embeddings.}
\label{section_multimodal_embedding}
\vic{For an entity $E_e$, we treat its textual description $T_e$ and visual context $V_e$ as a text-image pair $P_e=\{T_e; V_e\}$. We similarly obtain the text-image pair representation of a mention: $P_m=\{T_m; V_m\}$. As  feature extractor in $P_e$ and $P_m$, we utilize a pre-trained CLIP model~\citep{Alec_2021}, which trains two neural-network-based encoders using a contrastive loss to match pairs of texts and images. More precisely,  take $P_e$ as an example, we obtain textual and visual embedding representations as follows:}


\smallskip
\noindent\textbf{$\triangleright$ Textual Modal Embedding.}
\vic{We first concatenate the entity name $N_e$ and  the corresponding textual description $T_e$ using the  special tokens [\texttt{EOT}] and [\texttt{SEP}] to obtain the input sequence \{[\texttt{EOT}]$N_e$ [\texttt{SEP}]$T_e$[\texttt{SEP}]\}. Then, we tokenize the input sequence to the token sequence $\{t_e^{[\texttt{EOT}]}, t_e^1; t_e^2; \ldots ;t_e^{l_e}\}$, where $l_e$+1 is the length of the token sequence. Additionally, we use CLIP's text encoder to extract the hidden states: $\{{\textit{\textbf{t}}_e^{[\texttt{EOT}]}; \textit{\textbf{t}}_e^1; \textit{\textbf{t}}_e^2; \ldots ; \textit{\textbf{t}}_e^{l_e}}\} \in \mathbb{R}^{(l_e+1)\times d_t}$, where $d_t$ is the dimension of textual features. Finally, we consider the embedding of [\texttt{EOT}] as the global textual feature $\textbf{T}_e^G \in \mathbb{R}^{d_t}$ and the entire hidden state embeddings as local textual features $\textbf{T}_e^L \in \mathbb{R}^{(l_e+1)\times d_t}$.}

\vic{Similarly, we can also obtain the global textual feature $\textbf{T}_m^G$ and local textual feature $\textbf{T}_m^L$ of the mention textual modality.}

\smallskip
\noindent\textbf{$\triangleright$ Visual Modal Embedding.}
\vic{Given an entity image $V_e \in \mathbb{R}^{C\times H\times W}$, we reshape $V_e$ into a sequence of image patches $\vartheta= \{v_e^1; v_e^2; \ldots;v_e^n\} \\ \in \mathbb{R}^{n\times (P^2 \cdot C)}$, where $H \times W$ is the original image resolution, $C$ is the number of channels, $P \times P$ represents the patch image resolution, $n=HW/P^2$ is the number of patches. We also prepend an additional [\texttt{CLS}] token to $\tau$ before the image patches to represent the visual global feature, the corresponding patches sequence becomes $\vartheta' =\{v_e^{[\texttt{CLS}]}; v_e^1; v_e^2;\ldots; v_e^n\}$. We feed $\vartheta'$ into the CLIP visual encoder and further apply a fully connected layer to convert the dimension of the output hidden status into $d_v$. The corresponding embedding is denoted as $\{\textit{\textbf{v}}_e^{[\texttt{CLS}]}; \textit{\textbf{v}}_e^1; \textit{\textbf{v}}_e^2; \ldots ; \textit{\textbf{v}}_e^n\} \in \mathbb{R}^{(n+1)\times d_v}$. We take the embedding of the special token [\texttt{CLS}] as the global feature $\textbf{V}_e^G \in \mathbb{R}^{d_v}$ and the full  embedding as local features $\textbf{V}_e^L \in \mathbb{R}^{(n+1)\times d_v}$, where $d_v$ is the dimension of the visual features. }

\vic{Similarly, we can also obtain the global visual feature $\textbf{V}_m^G$ and local visual feature $\textbf{V}_m^L$ of the mention visual modality.}

\begin{figure}[t!]
    \centering
    \includegraphics[width=0.48\textwidth]{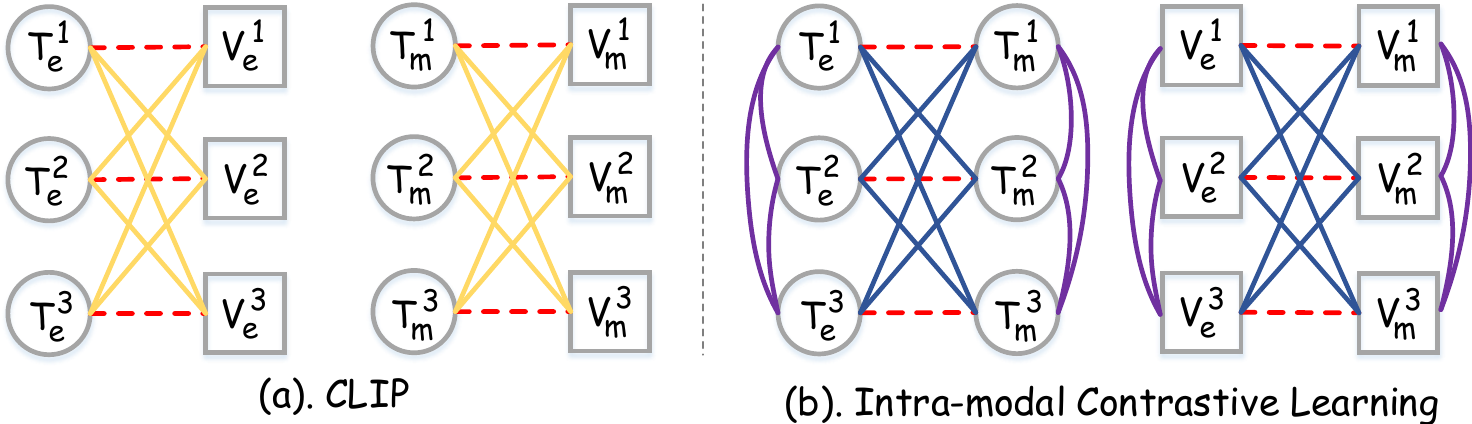}
    \caption{\vic{Illustrative comparison of intra-modal contrastive learning with CLIP, where the red dashed lines represent the positive samples, yellow lines denote the negative samples in CLIP, purple and blue lines represent the inner-source and intra-source negative samples. Circles and squares represent textual and visual features, $e$ and $m$ represent entity and mention, respectively.}}
    \label{figure_model_sub}
\end{figure}


\subsubsection{Intra-modal Contrastive Learning.}
\vic{CLIP learns a joint vision-language embedding space by bringing  matching  image-text representations together, while pushing unpaired instances away from each other. However, CLIP lacks an explicit mechanism that ensures that similar features from the same modality stay close  in the joint embedding~\citep{Jinyu_2022, Alex_2022, Mohammadreza_2021}. For example in Figure~\ref{figure_model_sub}, given the entity textual feature $\textbf{T}_e^{i}$, entity visual feature $\textbf{V}_e^{i}$, mention textual feature $\textbf{T}_m^{i}$ and mention visual feature $\textbf{V}_m^{i}$, CLIP only considers   negative samples of $\textbf{T}_e^{i}$ from different modalities (i.e., $\textbf{V}_e^{j}$ for $j{\neq} i$), ignoring the possibility of having negative samples from the same modality, i.e., $\textbf{T}_e^{j}$ and $\textbf{T}_m^{j}$. CLIP is able to map image-text pairs close together in the embedding space, while it fails to ensure that similar inputs from the same modality stay close by, which may result in degraded representations~\citep{Jinyu_2022, Mohammadreza_2021}. However, the main aim of multimodal entity linking is to retrieve the ground truth entity most relevant to the mention,  so 
simply considering the interaction between  different modalities will inevitably lead to a  one-sided embedding representation. Thus, it is necessary to consider the semantic difference between positive and negative samples within the same modality. To obtain better discriminative embedding representations that are faithful to unimodal differences, we introduce an \textbf{I}ntra-modal \textbf{C}ontrastive \textbf{L}earning module (\textbf{ICL}), composed of the following two steps: }

\smallskip
\noindent\textbf{$\triangleright$  Step 1. Obtaining  Positive and Negative Samples.} 
\vic{We take the textual modality as an example. Assume that there are $u$ entity textual descriptions $\{T_e^1, T_e^2, \ldots, T_e^u\}$ and $u$ mention textual descriptions $\{T_m^1, T_m^2, \ldots, T_m^u\}$, we can obtain the global textual feature corresponding to each piece of text based on the \textit{Textual Modal Embedding} in Section~\ref{section_multimodal_embedding}. Note that for  computational efficiency, we only perform intra-modal contrastive learning on the global feature level, we leave the interaction of local features to the \textit{Intra-modal Matching Network} in  Section~\ref{section_3_3}. Furthermore, we found out that introducing contrastive learning at  the local feature level  will actually lead to  performance degradation.  Therefore, we omit superscripts $G$ and $L$  from textual embeddings of entities and mentions and simply denote them as $\{\textbf{T}_e^{1}, \textbf{T}_e^{2}, \ldots, \textbf{T}_e^{u}\}$ and $\{\textbf{T}_m^{1}, \textbf{T}_m^{2}, \ldots, \textbf{T}_m^{u}\}$. For the entity embedding $\textbf{T}_e^{i}$, the matching mention embedding $\textbf{T}_m^{i}$ provides a positive sample, while the embeddings of other entities and mentions are  regarded as negative samples. More precisely, negative samples come from two sources: inner-source from the entity aspect and inter-source from the mention aspect. Inner-source means that the negative samples for $\textbf{T}_e^{i}$ are entity embeddings $\mathcal{N}_{e}=\{\textbf{T}_e^{j}~|~i {\neq} j\}$  in the same entity view where $\textbf{T}_e^{i}$ is located, while inter-source means that the negative samples are the embeddings $\mathcal{N}_{m}=\{\textbf{T}_m^{j}~|~i {\neq} j\}$ in the mention view where $\textbf{T}_e^{i}$ is not located.}

\smallskip
\noindent\textbf{$\triangleright$ Step 2.  Contrastive Learning Loss Computation.}
\vic{Take the entity textual embedding $\textbf{T}_e^{i}$ and the mention textual embedding $\textbf{T}_m^{i}$ as an example, we define the intra-modal contrastive learning loss of the positive pair ($\textbf{T}_e^{i}$, $\textbf{T}_m^{i}$,) as follows:}
\begin{equation}
\label{equation_2}
\left\{
\begin{array}{l}
\mathcal{L}(\textbf{T}_e^{i}, \textbf{T}_m^{i})=-{\rm log}\frac{\theta(\textbf{T}_e^{i}, \textbf{T}_m^{i})}{\theta(\textbf{T}_e^{i}, \textbf{T}_m^{i})+\beta\,\cdot\,\Phi_{inner}+\gamma\,\cdot\,\Phi_{inter}}\vspace{1.3ex}\\
\Phi_{inner}=\sum\limits_{\textbf{T}_e^{j} \in \mathcal{N}_{e}}\theta(\textbf{T}_e^{i}, \textbf{T}_e^{j})\\
\Phi_{inter}=\sum\limits_{\textbf{T}_m^{j} \in \mathcal{N}_{m}}\theta(\textbf{T}_e^{i}, \textbf{T}_m^{j})\\ 
\end{array} \right.
\end{equation}
\vic{where $\theta(x,y)=e^{\delta(x,y)/\tau}$, $\tau$ is a temperature parameter, $\delta(x,y)$ is the cosine similarity to measure the distance between two embeddings~\citep{Yanqiao_2021, Mohammadreza_2021, Zhiwei_2023}. The $\Phi_{inner}$ and $\Phi_{inter}$ in the denominator sum up the inner-source and inter-source intra-modal negative samples, respectively. $\beta$ and $\gamma$ are the hyper-parameters to control the inner-source and inter-source alignment importance, respectively. While the nominator is symmetric, the denominator is not, so for the positive pair  $(\textbf{T}_m^{i}, \textbf{T}_e^{i})$, the corresponding intra-modal loss is $\mathcal{L}(\textbf{T}_m^{i}, \textbf{T}_e^{i})$.} 

\vic{Similarly, for the visual positive sample pairs $(\textbf{V}_e^{i}, \textbf{V}_m^{i})$ and $(\textbf{V}_m^{i}, \textbf{V}_e^{i})$, the corresponding contrastive losses are $\mathcal{L}(\textbf{V}_e^{i}, \textbf{V}_m^{i})$ and $\mathcal{L}(\textbf{V}_m^{i}, \textbf{V}_e^{i})$. The final loss is the average of all textual and visual modalities pair losses, denoted as: }
\begin{equation}
\label{equation_3}
\mathcal{L}_{cl} = {\rm avg}(\sum\limits_{i}[\mathcal{L}(\textbf{T}_e^{i}, \textbf{T}_m^{i})+\mathcal{L}(\textbf{T}_m^{i}, \textbf{T}_e^{i})+\mathcal{L}
(\textbf{V}_e^{i}, \textbf{V}_m^{i})+\mathcal{L}(\textbf{V}_m^{i}, \textbf{V}_e^{i})])
\end{equation}

\subsection{IMN: Intra-modal Matching Network}
\vic{Through the CLIP encoder, we can obtain global and local features of textual and visual modalities.  Most previous works either exploit global features while overlooking the local features, or measure the local feature similarity whereas ignoring the global coherence~\citep{Ledell_2020}.  MIMIC~\citep{Pengfei_2023} considers the interaction  between global and local features but it uses  different independent  mechanisms for the textual and visual modalities, making the global and local interaction  deeply coupled with the modality. To alleviate the deep coupling between the interaction strategy and  the modality type, we introduce the \textbf{I}ntra-modal \textbf{M}atching \textbf{N}etwork (\textbf{IMN}) module to uniformly capture the interaction between local and global features within a modality.  IMN contains two sub-modules, Coarse-grained \textbf{G}lobal-to-\textbf{G}lobal matching (\textbf{G2G}) and Fine-grained \textbf{G}lobal-to-\textbf{L}ocal matching (\textbf{G2L}). Take the textual modality as an example, using the CLIP encoder, we can obtain the entity global textual feature $\textbf{T}_e^G$, entity local textual feature $\textbf{T}_e^L$, mention global textual feature $\textbf{T}_m^G$ and mention local textual feature $\textbf{T}_m^L$. We will use these features as the input to the  G2G and G2L sub-modules. }


\smallskip
\noindent\textbf{$\triangleright$ Coarse-grained Global-to-Global matching.} \vic{To measure  global consistency, we directly perform  the dot product $\odot$ between the entity global feature $\textbf{T}_e^G$ and the mention global feature $\textbf{T}_m^G$ to obtain the coarse-grained global-to-global matching score,  formulated as:}
\begin{equation}
\label{equation_4}
\mathcal{M}_T^{G2G}=\textbf{T}_e^G\odot\textbf{T}_m^G
\end{equation}

\smallskip
\noindent\textbf{$\triangleright$ Fine-grained Global-to-Local matching.} \vic{We introduce an attention mechanism to  explore the fine-grained clues among local features to obtain the fine-grained global-to-local matching score,  formulated as:}
\begin{equation}
\label{equation_5}
\left\{
\begin{array}{l}
Q, K, V = {\rm MLP}_1(\textbf{T}_e^L),\,{\rm MLP}_2(\textbf{T}_m^L),\,{\rm MLP}_3(\textbf{T}_m^L)\vspace{1.3ex}\\
\alpha^L = {\rm Mean}({\rm Softmax}(\frac{Q\otimes K^T}{\sqrt{d_s}})\otimes V) \\
\mathcal{M}_T^{G2L}=\textbf{T}_e^G\odot\alpha^L\\ 
\end{array} \right.
\end{equation}
\label{section_3_3}
\vic{where $\{{\rm MLP}_1, {\rm MLP}_2, {\rm MLP}_3\}$: $\mathbb{R}^{d_t} \rightarrow \mathbb{R}^{d_s}$ are three multi-layer perceptron networks, $d_s$ denotes the scaled dimension size, $\otimes$ is the matrix multiplication operation, ${\rm Mean}(\cdot)$ represents the mean pooling operation, $\alpha^T$ is the attention score generated from local features to impose constraints on global features.}

\vic{Afterwards, we average the global-to-global and global-to-local matching scores to obtain the textual intra-modal matching score $\mathcal{M}_T=(\mathcal{M}_T^{G2G}+\mathcal{M}_T^{G2L})/2$. }
\vic{Similarly, for the visual modality, we can obtain the global-to-global matching score $\mathcal{M}_V^{G2G}$ and global-to-local matching score $\mathcal{M}_V^{G2L}$, and  the final  combined  matching score $\mathcal{M}_V=(\mathcal{M}_V^{G2G}+\mathcal{M}_V^{G2L})/2$.}

\subsection{CMN: Cross-modal Matching Network}
\label{section_cmn}
\vic{Since the embeddings of different modalities are separately matched in the  IMN module, it is difficult to model the complex interaction between modalities solely based on an intra-modal matching mechanism. Although the CLIP model considers the alignment of information between multi-modalities, it  is not specifically tailored for the MEL task. Therefore, it is necessary to appropriately adapt the inter-modal interaction to the embedding representation obtained from CLIP, based on the characteristics of the MEL task. To reduce the gap between the distribution over different modalities, we introduce the \textbf{C}ross-modal \textbf{M}atching \textbf{N}etwork (\textbf{CMN}) module, which contains two-way matching strategies: \textbf{T}extual-to-\textbf{V}isual matching (T2V) and \textbf{V}isual-to-\textbf{T}extual matching (V2T). Considering that the difference between T2V and V2T is only in the input embeddings content, we mainly give details for T2V. The T2V strategy  has as input  an entity-level textual-to-visual matching between the entity global textual feature $\textbf{T}_e^G$ and entity local visual features $\textbf{V}_e^L$, and a mention-level textual-to-visual matching between the mention global textual feature $\textbf{T}_m^G$ and the mention local visual features $\textbf{V}_m^L$. The entity-level and mention-level textual-to-visual matching mechanisms both include the following three steps. We  use the entity-level  to explain the details.}
\begin{enumerate}[itemsep=0.5ex, leftmargin=5mm]
\item \textit{Bidirectional Matching Interaction.} 
\vic{First, we employ a one-layer MLP to scale the dimensions of $\textbf{T}_e^G$ and $\textbf{V}_e^L$ to the same size. Then, we introduce an attention mechanism to obtain the local visual-aware global attention $\textbf{H}_e^{\,GL}$ and global textual-aware local attention $\textbf{H}_e^{\,LG}$. Finally, we utilize the attention content to get an enhanced entity global textual feature $\textbf{T}_e^{\overline{G'}}\in\mathbb{R}^{d_c}$ and local visual features $\textbf{V}_e^{\overline{L'}}\in\mathbb{R}^{(n+1)\times d_c}$. 
The entire interaction process is bidirectional since the data flows from local to global and global to local.}
\begin{equation}
\label{equation_6}
\left\{
\begin{array}{l}
\textbf{T}_e^{\overline{G}}={\rm MLP_4}(\textbf{T}_e^G),\,\,\textbf{V}_e^{\overline{L}}={\rm MLP_5}(\textbf{V}_e^L)\vspace{1.3ex}\\
\textbf{H}_e^{\,GL}={\rm Softmax}(\textbf{T}_e^{\overline{G}}\otimes\textbf{V}_e^{\overline{L}}),\,\,\textbf{H}_e^{\,LG}={\rm Softmax}(\textbf{V}_e^{\overline{L}}\otimes\textbf{T}_e^{\overline{G}})\vspace{1.3ex} \\
\textbf{T}_e^{\overline{G'}}={\rm Relu}(\textbf{H}_e^{\,GL}\otimes\textbf{V}_e^{\overline{L}}),\,\,\textbf{V}_e^{\overline{L'}}={\rm Relu}(\textbf{H}_e^{\,LG}\otimes\textbf{T}_e^{\overline{G}}) \\
\end{array} \right.
\end{equation}
\vic{where ${\rm MLP_4}: \mathbb{R}^{d_t} \rightarrow \mathbb{R}^{d_c}$ and  ${\rm MLP_5}: \mathbb{R}^{d_v} \rightarrow \mathbb{R}^{d_c}$ are  multi-layer perceptron networks, $d_c$ represents the scaled embedding dimension, ${\rm Softmax}(\circ)$ and ${\rm Relu}(\circ)$ are two different activation functions.} To better understand Equation~\ref{equation_6}, we elaborate on the dimensional transformation occurring in it. It should be noted that to simplify the presentation, all dimensions do not include the batch size dimension, because the batch size of all operations is the same. (\textit{i}): for the first row, we change $\textbf{T}_e^G\in\mathbb{R}^{d_t}$ and $\textbf{V}_e^L\in\mathbb{R}^{(n+1)\times d_v}$ to $\textbf{T}_e^{\overline{G}}\in\mathbb{R}^{d_c}$ and $\textbf{V}_e^{\overline{L}}\in\mathbb{R}^{(n+1)\times d_c}$; (\textit{ii}): for the second row, we first extend the first dimension of $\textbf{T}_e^{\overline{G}}$ from $\mathbb{R}^{d_c}$ to $\mathbb{R}^{1\times d_c}$, then we perform a matrix multiplication operation on the expanded $\textbf{T}_e^{\overline{G}}$ and the transpose of $\textbf{V}_e^{\overline{L}}$, and following the Softmax activation function to obtain $\textbf{H}_e^{\,GL}\in\mathbb{R}^{1\times (n+1)}$. Similarly, for $\textbf{H}_e^{\,LG}\in\mathbb{R}^{(n+1)\times 1}$; (\textit{iii}): for the third row, we perform a matrix multiplication operation on $\textbf{H}_e^{\,GL}\in\mathbb{R}^{1\times (n+1)}$ and $\textbf{V}_e^{\overline{L}}\in\mathbb{R}^{(n+1)\times d_c}$ to get $\textbf{T}_e^{\overline{G'}}\in\mathbb{R}^{1\times d_c}$ (for simplicity, we can omit the 1 in the dimension). Similarly, for $\textbf{V}_e^{\overline{L'}}\in\mathbb{R}^{(n+1)\times d_c}$.

\item \textit{Multi-head Attention Fusion.} \vic{
\vic{Given the enhanced entity local visual feature representation $\textbf{V}_e^{\overline{L'}}$
and the enhanced entity global textual embedding $\textbf{T}_e^{\overline{G'}}$, 
we first concatenate them  and  then introduce a multi-head attention mechanism to convert the concatenated vector $\textbf{h}_e^{T2V}$ into $\textbf{H}_e^{T2V}\in\mathbb{R}^{d_c}$.} Additionally, we apply a gate operation to fuse  $\textbf{H}_e^{T2V}$ and the original entity global textual feature $\textbf{T}_e^G$ to obtain the entity-level textual-to-visual matching representation $\mathcal{M}_e^{T2V}$. More precisely, this process is formulated as follows: }
\begin{equation}
\label{equation_7}
\left\{
\begin{array}{l}
\textbf{h}_e^{T2V}={\rm Concat}[\textbf{V}_e^{\overline{L'}}; \textbf{T}_e^{\overline{G'}}]
\vspace{1.3ex}\\
\textbf{H}_e^{T2V}={\rm avg}(\sum_{k=1}^{K}({\rm Softmax}(\omega_k\circledast\textbf{h}_e^{T2V})\circledast\textbf{h}_e^{T2V}))
\vspace{1.3ex} \\
\mathcal{M}_e^{T2V}=\textbf{H}_e^{T2V}\oplus{\rm Tanh}(\textbf{T}_e^G)\circledast\textbf{T}_e^G\\
\end{array} \right.
\end{equation}
\vic{where ${\rm Concat}[\circ;\circ]$ and ${\rm avg}(\circ)$ denote the concatenation and average operation, respectively. ${\rm Tanh}(\circ)$ is an activation function, $K$ represents the number of heads, $\oplus$ denotes the element-wise addition, $\circledast$ is  the algebraic multiplication operation, and $\omega_k>0$ is the temperature controlling the sharpness of scores. It should be noted that before executing the ${\rm Concat}$ operation, the dimensions of $\textbf{T}_e^{\overline{G'}}$ need to be expanded to convert it from $\textbf{T}_e^{\overline{G'}}\in\mathbb{R}^{d_c}$ to $\textbf{T}_e^{\overline{G'}}\in\mathbb{R}^{1\times d_c}$.}
\item 
\vic{Analogously, taking the mention global textual feature $\textbf{T}_m^G$ and mention local visual features $\textbf{V}_m^L$ as inputs, after going through Step 1 and Step 2 above, we obtain the mention-level textual-to-visual matching representation $\mathcal{M}_m^{T2V}$. The textual-to-visual matching score $\mathcal{M}^{T2V}$ is then  calculated  as follows:}
\begin{equation}
\label{equation_8}
\mathcal{M}^{T2V}=\mathcal{M}_e^{T2V}\odot\mathcal{M}_m^{T2V}
\end{equation}

\vic{In the same way, we can obtain the entity-level visual-to-textual matching representation $\mathcal{M}_e^{V2T}$ and the mention-level visual-to-textual matching embedding $\mathcal{M}_m^{V2T}$. We can also apply the dot product of $\mathcal{M}_e^{V2T}$ and $\mathcal{M}_m^{V2T}$ to get the visual-to-textual matching score $\mathcal{M}^{V2T}$. Finally, we combine the textual-to-visual score and the visual-to-textual score to get  the cross-modal matching network result $\mathcal{M}_C = (\mathcal{M}_m^{T2V}+\mathcal{M}_m^{V2T})/2$.}
\end{enumerate}

\subsection{Joint Training}
\vic{Consider the textual intra-modal matching score $\mathcal{M}_T$ and the visual intra-modal matching score $\mathcal{M}_V$ from the IMN module and the cross-modal matching score $\mathcal{M}_C$ from the CMN module. We can obtain the union matching score as $\mathcal{M}_U=(\mathcal{M}_T+\mathcal{M}_V+\mathcal{M}_C)/3$. We select the Unit-Consistent Objective Function~\citep{Pengfei_2023} $\mathcal{L}_{uco}$ as the basic loss function. Specifically, the entire joint training process loss $\mathcal{L}_{Joint}$ contains three aspects of loss contents as shown in Equation~\ref{equation_9}: \textit{i}) the intra-modal contrastive learning loss $\mathcal{L}_{cl}$; \textit{ii}) the union matching loss $\mathcal{L}_U=\mathcal{L}_{uco}(\mathcal{M}_U)$; \textit{iii}) to avoid the whole model to excessively rely on the union score, we also introduce the independent intra-model textual loss $\mathcal{L}_T=\mathcal{L}_{uco}(\mathcal{M}_T)$, the intra-modal visual loss $\mathcal{L}_V=\mathcal{L}_{uco}(\mathcal{M}_V)$, and the cross-modal loss $\mathcal{L}_C=\mathcal{L}_{uco}(\mathcal{M}_C)$.}
\begin{equation}
\label{equation_9}
\mathcal{L}_{Joint}=\mathcal{L}_{cl}+\mathcal{L}_U+\mathcal{L}_T+\mathcal{L}_V+\mathcal{L}_C
\end{equation}

\section{Experiments}
To evaluate the effectiveness of ${\rm M^3EL}$, we conduct the following four  experiments: including main results, ablation studies, parameter sensitivity (see \textcolor{blue}{\textbf{Appendix \hyperlink{parameter_sensitivity}{A}}}), and additional experiments (see \textcolor{blue}{\textbf{Appendix \hyperlink{additional_experiments}{B}}}).

\subsection{Experimental Setup}

\smallskip
\noindent\textbf{$\triangleright$ Datasets.}
\vic{We evaluate  the ${\rm M^3EL}$ model  on three well-known datasets: WikiMEL~\citep{Peng_2022}, RichpediaMEL~\citep{Peng_2022}, and WikiDiverse~\citep{Xuwu_2022}.  WikiMEL has more than 22K multimodal samples, where entities come from WikiData~\citep{Denny_2014} and their corresponding textual and visual descriptions from  WikiPedia. RichpediaMEL has more than 17K multimodal samples, where the entity ids are from the large-scale multimodal knowledge graph Richpedia~\citep{Meng_2020} and the corresponding multimodal information from WikiPedia. WikiDiverse has more than 7k multimodal samples,  constructed from WikiNews. For fair comparison, we adopt the dataset division ratio from MIMIC~\citep{Pengfei_2023}. The statistics of these datasets are  shown in  Table~\ref{table_statistics_datasets}.}

\begin{table}[!htp]
\setlength{\abovecaptionskip}{0.03cm} 
\renewcommand\arraystretch{1.2}
\setlength{\tabcolsep}{0.60em}
\centering
\small
\caption{Statistics of three different datasets. "Num." and "Img." denote number and image, respectively.}
\begin{tabular*}{0.98\linewidth}{@{}c|ccc@{}}
\hline
\multicolumn{1}{c|}{\textbf{Statistics}} &\multicolumn{1}{c}{\textbf{WikiMEL}} &\multicolumn{1}{c}{\textbf{RichpediaMEL}} &\multicolumn{1}{c}{\textbf{WikiDiverse}}\\
\hline
\# Num. of sentences  &22,070  &17,724  &7,405 \\
\# Num. of entities  &109,976  &160,935  &132,460 \\
\# entities with Img. &67,195  &86,769  &67,309 \\
\# Num. of mentions  &25,846  &17,805  &15,093 \\
\# Img. of mentions   &22,136  &15,853  &6,697 \\
\# mentions in train  &18,092  &12,463  &11,351 \\
\# mentions in valid  &2,585  &1,780  &1,664 \\
\# mentions in test  &5,169  &3,562  &2,078 \\
\hline
\end{tabular*}
\label{table_statistics_datasets}
\end{table}

\smallskip
\noindent\textbf{$\triangleright$ Evaluation Metrics.} 
\vic{We evaluate the  performance using two  common metrics: MRR and Hits@\textit{N}. MRR defines the inverse of the rank for the first correct answer, Hits@\textit{N} is the proportion of correct answers ranked in the top \textit{N}, with \textit{N}=\{1,3,5\}. The higher the values of MRR or Hits@\textit{N}, the better the performance.}

\smallskip
\noindent\textbf{$\triangleright$ Implementation Details.} 
\vic{We conduct all experiments on six 32G Tesla V100 GPUs, and use the AdamW~\citep{Ilya_2019} optimizer. Following MIMIC~\citep{Pengfei_2023}, we employ the pre-trained CLIP-Vit-Base-Patch32 model\footnote{https://huggingface.co/openai/clip-vit-base-patch32} as the initialization textual and visual encoder.
For the textual modality, we set the maximal input length of the text to 40 and the dimension of the textual output features $d_t$  to 512. For the visual modality, we rescale all the images into a 224$\times$224 resolution and set the dimension of the visual output features $d_v$  to 96. The scaled dimension size $d_s$ is set to 96 and the patch size $P$ is set to 32.
We use  grid search to select the optimal hyperparameters, mainly including: the learning rate $lr\in$ \{5e-6, \textbf{1e-5}, 2e-5, 3e-5, 4e-5\}, the batch size $bs\in$ \{48, 64, 80, \textbf{96}, 112\}, the temperature coefficient $\tau\in$ \{\textbf{0.03}, 0.10, 0.25, 0.5, 0.75\} in Equation~\ref{equation_2}, the number of heads $K\in$ \{3, 4, \textbf{5}, 6, 7\}, the inner-source alignment weight $\beta\in$ \{0.2, 0.4, 0.6, \textbf{0.8}, 1.0\}, the inter-source alignment weight $\gamma\in$ \{0.2, 0.4, 0.6, 0.8, \textbf{1.0}\}.
}

\smallskip
\noindent\textbf{$\triangleright$ Baselines.} \vic{We compare ${\rm M^3EL}$ with three types of baselines. (\textit{i}) textual-based methods, including BLINK~\citep{Ledell_2020}, BERT~\citep{Jacob_2019} and RoBERTa~\citep{Yinhan_2019}; (\textit{ii}) representation learning frameworks, including DZMNED~\citep{Seungwhan_2018}, JMEL~\citep{Omar_2020}, VELML~\citep{Qiushuo_2022} and GHMFC~\cite{Peng_2022}; (\textit{iii}) vision-and-language-based methods, including CLIP~\citep{Alec_2021}, ViLT~\citep{Wonjae_2021}, ALBEF~\citep{Junnan_2021}, METER~\citep{Zi_2022} and MIMIC~\citep{Pengfei_2023}; (\textit{iv}) generative-based methods, including GPT-3.5-Turbo~\citep{OpenAI_2023} and GEMEL~\citep{Senbao_2024}.}



\subsection{Main Results}
\vic{We conduct experiments on the WikiMEL, RichpediaMEL and WikiDiverse datasets in the normal and low resource settings. The corresponding results are shown in Tables~\ref{table_main_results} and~\ref{table_low_resource}.}

\begin{table*}[!htp]
\setlength{\abovecaptionskip}{0.18cm}
\renewcommand\arraystretch{1.3}
\setlength{\tabcolsep}{0.36em}
\centering
\small
\caption{\vic{Evaluation of different models on three MEL datasets, all the baselines results are from the MIMIC~\citep{Pengfei_2023} paper. Best scores are highlighted in \textbf{bold}, the second best scores are \uline{underlined}. It should be noted that there are a large number of - labeled results in GPT-3.5-Turbo and GEMEL. The main reasons are, on the one hand, \citep{Senbao_2024} only provides  results for the  WikiMEL and WikiDiverse datasets, and the WikiDiverse dataset is completely different from that used by other baselines.  \citep{Senbao_2024} only uses Hits@1 as the evaluation indicator.}}
\begin{tabular*}{0.875\linewidth}{@{}ccccccccccccc@{}}
\bottomrule
\multicolumn{1}{c}{\multirow{2}{*}{\textbf{Methods}}} & \multicolumn{4}{c}{\textbf{WikiMEL}} & \multicolumn{4}{c}{\textbf{RichpediaMEL}} & \multicolumn{4}{c}{\textbf{WikiDiverse}}\\
\cline{2-5}\cline{6-9}\cline{10-13}
& \textbf{MRR} & \textbf{Hits@1} & \textbf{Hits@3} & \textbf{Hits@5} & \textbf{MRR} & \textbf{Hits@1} & \textbf{Hits@3} & \textbf{Hits@5} & \textbf{MRR}  & \textbf{Hits@1} & \textbf{Hits@3} & \textbf{Hits@5} \\
\hline
BLINK~\citep{Ledell_2020}   &81.72  &74.66  &86.63  &90.57   &71.39 &58.47  &81.51  &88.09  &69.15  &57.14  &78.04  &85.32 \\ 
BERT~\citep{Jacob_2019}   &81.78  &74.82  &86.79  &90.47   &71.67 &59.55  &81.12  &87.16  &67.38  &55.77  &75.73  &83.11 \\ 
RoBERTa~\citep{Yinhan_2019}   &80.86  &73.75  &85.85  &89.80   &72.80 &61.34  &81.56  &87.15  &70.52  &59.46  &78.54  &85.08 \\ 
\hline
DZMNED~\citep{Seungwhan_2018}   &84.97  &78.82  &90.02  &92.62   &76.63 &68.16  &82.94  &87.33  &67.59  &56.90  &75.34  &81.41 \\ 
JMEL~\citep{Omar_2020}   &73.39  &64.65  &79.99  &84.34   &60.06 &48.82  &66.77  &73.99  & 48.19  &37.38  &54.23  &61.00 \\ 
VELML~\citep{Qiushuo_2022}   &83.42  &76.62  &88.75  &91.96   &77.19 &67.71  &84.57  &89.17  & 66.13  &54.56  &74.43  &81.15 \\ 
GHMFC~\cite{Peng_2022}   &83.36  &76.55  &88.40  &92.01   &80.76 &72.92  &86.85  &90.60  & 70.99  &60.27  &79.40  &84.74 \\ 
\hline
CLIP~\citep{Alec_2021}   &88.23  &83.23  &92.10  &94.51   &77.57 &67.78  &85.22  &90.04  &
71.69  &61.21  &79.63  &85.18 \\ 
ViLT~\citep{Wonjae_2021}   &79.46  &72.64  &84.51  &87.86   &56.63 &45.85  &62.96  &69.80  &45.22  &34.39  &51.07  &57.83 \\ 
ALBEF~\citep{Junnan_2021}   &84.56  &78.64  &88.93  &91.75   &75.29 &65.17  &82.84  &88.28  &69.93  &60.59  &75.59  &81.30 \\ 
METER~\citep{Zi_2022}   &79.49  &72.46  &84.41  &88.17   &74.15 &63.96  &82.24  &87.08  &63.71  &53.14  &70.93  &77.59 \\ 
MIMIC~\citep{Pengfei_2023}   &\uline{91.82}  &87.98  &95.07  &96.37   &86.95 &81.02  &91.77  &94.38  &73.44  &63.51  &81.04  &86.43 \\ 
\hline
GPT-3.5-Turbo~\citep{OpenAI_2023}   &-  &73.80  &-  &-   &- &-  &-  &-  &-  &-  &-  &- \\ 
GEMEL~\citep{Senbao_2024}   &-  &82.60  &-  &-   &- &-  &-  &-  &-  &-  &-  &- \\ 
\hline
${\rm M^3EL}_{\rm attr}$   &\textbf{92.30}  &\uline{88.49}  &\textbf{95.55}  &\textbf{97.10}   &\uline{88.08}    &\uline{82.79} &\uline{92.53}  &\uline{94.69}  &\uline{77.26}  &\uline{68.77}  &\uline{83.59}  &\uline{87.58} \\ 
${\rm M^3EL}_{\rm desc}$   &\textbf{92.30}  &\textbf{88.84}  &\uline{95.20}  &\uline{96.71}   &\textbf{88.26} &\textbf{82.82} &\textbf{92.73}  &\textbf{95.34}  &\textbf{81.29}  &\textbf{74.06}  &\textbf{86.57}  &\textbf{90.04} \\
\hline
\end{tabular*}
\label{table_main_results}
\end{table*}

\begin{table*}[!htp]
\setlength{\abovecaptionskip}{0.18cm}
\renewcommand\arraystretch{1.3}
\setlength{\tabcolsep}{0.36em}
\centering
\small
\caption{\vic{Evaluation of different models in low resource settings on four datasets. All the baselines results are from the MIMIC~\citep{Pengfei_2023} paper. Best scores are highlighted in \textbf{bold}, the second best scores are \uline{underlined}. H@\textit{i} is the abbreviation of Hits@\textit{i}.}}
\begin{tabular*}{0.90\linewidth}{@{}ccccccccccccccccc@{}}
\bottomrule
\multicolumn{1}{c}{\multirow{2}{*}{\textbf{Methods}}} & \multicolumn{4}{c}{\textbf{RichpediaMEL(10\%)}} & \multicolumn{4}{c}{\textbf{RichpediaMEL(20\%)}} & \multicolumn{4}{c}{\textbf{WikiDiverse(10\%)}} & \multicolumn{4}{c}{\textbf{WikiDiverse(20\%)}}\\
\cline{2-5}\cline{6-9}\cline{10-13}\cline{11-17}
& \textbf{MRR} & \textbf{H@1} & \textbf{H@3} & \textbf{H@5} & \textbf{MRR} & \textbf{H@1} & \textbf{H@3} & \textbf{H@5} & \textbf{MRR}  & \textbf{H@1} & \textbf{H@3} & \textbf{H@5} &
\textbf{MRR}  & \textbf{H@1} & \textbf{H@3} & \textbf{H@5} \\
\hline
DZMNED~\citep{Seungwhan_2018}  &31.79 	&22.57 	&34.95 	&41.33  &47.01 	&36.38 	&52.25  &58.28   &19.99 	&11.45 	&22.52 	&29.50  &40.97 	&28.73   &47.35    &56.69 \\
JMEL~\citep{Omar_2020}   &25.01 	&16.70 	&27.68 	&33.63   &39.38  &28.92	&43.35 	&50.59  &28.26 	&19.97 	&32.19 	&37.58  &39.05 	&29.26 	&44.23 	&49.90 \\
VELML~\citep{Qiushuo_2022}   &35.52 	&27.15 	&38.60 	&43.99  &59.24 	&48.85 	&64.91 	&71.76  &40.70 	&30.51 	&46.20 	&52.36  &54.76 	&43.65 	&61.36 	&67.66 \\
GHMFC~\cite{Peng_2022}   &76.69 	&\uline{68.00} 	&83.38 	&87.73  &80.42 	&72.57 	&86.69 	&90.15  &59.56 	&48.08 	&66.31 	&74.25  &63.46 	&51.73 	&71.85 	&78.54 \\
\hline
CLIP~\citep{Alec_2021}    &72.51 	&62.66 	&79.14 	&85.06  &73.72 	&64.32 	&79.59 	&85.54  &69.49 	&59.87 	&76.52 	&81.57  &69.95 	&59.96 	&77.05 	&82.24 \\
ViLT~\citep{Wonjae_2021}    &17.05 	&11.73 	&18.59 	&22.07  &38.81 	&30.24 	&42.39 	&48.40  &19.57 	&13.19 	&21.27 	&26.37  &29.48 	&20.93 	&32.92 	&38.93 \\
ALBEF~\citep{Junnan_2021}   &72.51 	&63.19 	&79.31 	&84.25  &73.02 	&64.21 	&79.47 	&85.32  &62.26 	&51.83 	&69.20 	&74.64  &66.56 	&56.40 	&73.87 	&78.97 \\
METER~\citep{Zi_2022}   &71.40 	&60.89 	&79.23 	&84.78  &71.82 	&61.51 	&79.56 	&84.48  &53.53 	&40.42 	&61.31 	&70.26  &53.46 	&40.23 	&61.16 	&70.45 \\
MIMIC~\citep{Pengfei_2023}   &74.62 	&64.49 	&82.03 	&87.59  &\textbf{82.73} 	&\textbf{75.60} 	&\textbf{88.63} 	&\uline{91.72}  &69.70 	&60.54 	&76.18 	&81.33  &63.46 	&61.01 	&77.67 	&83.35 \\
\hline
${\rm M^3EL}_{\rm attr}$   &\uline{77.24} 	&67.41 	&\uline{84.87} 	&\uline{89.25}  &81.53 	&73.33 	&88.01 	&91.61  &\uline{73.80} 	&\uline{65.88} 	&\uline{78.83} 	&\uline{83.73}  &\uline{75.23} 	&\uline{67.08} 	&\uline{80.56} 	&\uline{85.32} \\
${\rm M^3EL}_{\rm desc}$   &\textbf{78.68} 	&\textbf{69.54} 	&\textbf{85.77} 	&\textbf{89.70}  &\uline{82.04} 	&\uline{74.20} 	&\uline{88.32} 	&\textbf{91.89}  &\textbf{74.61} 	&\textbf{66.36} 	&\textbf{80.70} 	&\textbf{84.31}  &\textbf{77.90} 	&\textbf{70.36} 	&\textbf{83.16} 	&\textbf{87.63} \\
\hline
\end{tabular*}
\label{table_low_resource}
\end{table*}

\smallskip
\noindent\textbf{$\triangleright$ Overall Performance Comparison.} \vic{Table~\ref{table_main_results} presents  results of the performance of  ${\rm M^3EL}$ and the baselines on the WikiMEL, RichpediaMEL and WikiDiverse datasets. \vic{We consider two variants of ${\rm M^3EL}$, ${\rm M^3EL}_{\rm attr}$ and ${\rm M^3EL}_{\rm desc}$, which differ in the available textual knowledge. For ${\rm M^3EL}_{\rm attr}$, following MIMIC,  we use the entity attribute knowledge as the textual  content of the corresponding entity. For ${\rm M^3EL}_{\rm desc}$, we use the textual description of  the entity in WikiData\footnote{https://query.wikidata.org/} as its textual modal content. The main motivation for  replacing attribute knowledge with the description knowledge from Wikidata is that the word length of the description  is larger than that of the textual attribute,  providing richer textual modal content. 
Under the same experimental conditions,} we observe that \zwadd{${\rm M^3EL}_{\rm attr}$} outperforms existing SoTA baselines by a large margin across all metrics. We have three main findings: (1) Compared with the  best performing baseline, MIMIC, \zwadd{${\rm M^3EL}_{\rm attr}$} respectively achieves 0.48\% and \zwadd{0.51\%} improvements in MRR and Hits@1  on the WikiMEL dataset. A similar behavior is observed  on the RichpediaMEL and WikiDiverse datasets. (2) On the three datasets, when compared with MIMIC, the improvement on the Hits@1 metric is higher than that on the Hits@3 metric, which shows that our \zwadd{${\rm M^3EL}_{\rm attr}$} pays more attention to highly accurate matching in the multimodal entity linking task. 
\vic{(3) The use of  description knowledge further improves the performance: on the three datasets, the performance of ${\rm M^3EL}_{\rm desc}$ is better than that of ${\rm M^3EL}_{\rm attr}$. The bigger gain is in  the WikiDiverse dataset. The main reason is that in the WikiDiverse dataset, a large percentage of the entities have short text descriptions, with the average text word length being only 1.24. After replacing attribute knowledge with description knowledge, the average word length of the entity text corresponding to the mention becomes 4.50. Therefore, we believe that the availability of textual knowledge is one of the important factors affecting the model performance. If there is no special explanation, ${\rm M^3EL}$ appearing in subsequent parts will uniformly refer to ${\rm M^3EL}_{\rm desc}$.}}

\smallskip
\noindent\textbf{$\triangleright$ Low Resource Setting.} \vic{To better understand the performance of ${\rm M^3EL}$l and existing baselines in low-resource scenarios, we conducted experiments on the RichpediaMEL and WikiDiverse datasets with 10\% and 20\% of the training data,  while keeping the validation and test sets unchanged. The corresponding results are presented  in Table~\ref{table_low_resource}. We  observe that ${\rm M^3EL}$ consistently achieves optimal performance on almost all subsets. We have the following two main findings. 1) Except for the case when only 20\% of the RichpediaMEL training data is used, ${\rm M^3EL}$ achieves optimal performance. Note that there is no unique best performing (second overall) baseline, switching between GHMFC, CLIP and MIMIC. This phenomenon reveals that ${\rm M^3EL}$ is more  stable and can adapt to a variety of datasets under different conditions. In fact, in the RichpediaMEL (20\%) scenario, ${\rm M^3EL}$ also achieves competitive results. 2) On the  WikiDiverse dataset as the training proportion increases, the gap between ${\rm M^3EL}$ and MIMIC gradually becomes larger. We attribute this performance improvement to the fact that the intra-modal contrastive learning module is able to obtain better discriminative representations as the size of the training data increases. The additional support of the multi-level matching mechanism of intra-modal and inter-modal has further brought a positive impact of final results.}
\subsection{Ablation Studies}
\vic{We conduct ablation experiments on the WikiMEL, RichpediaMEL and WikiDiverse datasets under different conditions.  Table~\ref{table_ablation_studies} presents results showing  the contribution of each component of ${\rm M^3EL}$. These include the following three points: \textit{a}) removing one of  $\mathcal{L}_U$, $\mathcal{L}_T$, $\mathcal{L}_V$, $\mathcal{L}_C$ or  $\mathcal{L}_{cl}$ from the loss function $\mathcal{L}_{Joint}$; \textit{b}) removing the textual intra-modal matching module (w/o $\mathcal{L}_T+\mathcal{M}_T$), the visual intra-modal matching module (w/o $\mathcal{L}_V+\mathcal{M}_V$), or the cross-modal matching module (w/o $\mathcal{L}_C+\mathcal{M}_C$) from ${\rm M^3EL}$, it should be noted that there is no ablation scenario w/o $\mathcal{M}_T$, w/o $\mathcal{M}_V$, w/o $\mathcal{M}_C$, because $\mathcal{L}_T$, $\mathcal{L}_V$, $\mathcal{L}_C$ are calculated based on $\mathcal{M}_T$, $\mathcal{M}_V$, $\mathcal{M}_C$, and removing $\mathcal{M}_T$, $\mathcal{L}_V$, $\mathcal{L}_C$ would mean that $\mathcal{L}_T$, $\mathcal{L}_V$, $\mathcal{L}_C$ are also removed; \textit{c}) replacing ICL with two contrastive learning losses, i.e., w/ InfoNCE~\citep{Kaiming_2020} and w/ MCLET~\citep{Zhiwei_2023}. In addition, we also conduct experiments Effect of Different Pooling Operation in Equation 5 and Resource Consumption Using Various Contrastive Loss.}


\begin{table*}[!htp]
\setlength{\abovecaptionskip}{0.05cm}
\renewcommand\arraystretch{1.3}
\setlength{\tabcolsep}{0.36em}
\centering
\small
\caption{Evaluation of ablation studies on three MEL datasets. Best scores are highlighted in \textbf{bold}, the second best scores are \uline{underlined}.}
\begin{tabular*}{0.86\linewidth}{@{}ccccccccccccc@{}}
\bottomrule
\multicolumn{1}{c}{\multirow{2}{*}{\textbf{Methods}}} & \multicolumn{4}{c}{\textbf{WikiMEL}} & \multicolumn{4}{c}{\textbf{RichpediaMEL}} & \multicolumn{4}{c}{\textbf{WikiDiverse}}\\
\cline{2-5}\cline{6-9}\cline{10-13}
& \textbf{MRR} & \textbf{Hits@1} & \textbf{Hits@3} & \textbf{Hits@5} & \textbf{MRR} & \textbf{Hits@1} & \textbf{Hits@3} & \textbf{Hits@5} & \textbf{MRR}  & \textbf{Hits@1} & \textbf{Hits@3} & \textbf{Hits@5} \\
\hline
w/o $\mathcal{L}_U$   &91.62 	&87.81 	&94.53 	&96.40 	&\textbf{88.34} 	&\uline{83.27} 	&92.59 	&94.50 	&75.46 	&66.12 	&82.44 	&87.15  \\ 
w/o $\mathcal{L}_T$   &91.64 	&87.60 	&94.97 	&96.67 	&85.62 	&78.66 	&91.58 	&94.41 	&80.63 	&\uline{73.68} 	&86.00 	&88.74  \\ 
w/o $\mathcal{L}_V$  &\uline{92.06} 	&\uline{88.37} 	&\textbf{95.20} 	&96.54 	&85.27 	&78.94 	&90.40 	&93.12 	&78.00 	&69.83 	&83.83 	&88.50  \\
w/o $\mathcal{L}_C$  &90.91 	&86.79 	&94.22 	&96.19 	&86.55 	&80.32 	&91.72 	&94.53 	&\uline{80.76} 	&73.34 	&\textbf{86.86} 	&\uline{89.85}  \\
w/o $\mathcal{L}_{cl}$	&91.89 	&88.35 	&94.74 	&96.42 	&88.23 	&\textbf{83.63} 	&91.66 	&94.08 	&75.66 	&66.31 	&82.68 	&87.82  \\
\hline
w/o $\mathcal{L}_T$+$\mathcal{M}_T$	&90.22 	&86.40 	&93.05 	&95.03 	&84.71 	&78.38 	&89.64 	&92.56 	&67.46 	&59.43 	&73.10 	&77.19   \\
w/o $\mathcal{L}_V$+$\mathcal{M}_V$	&87.37 	&81.72 	&91.78 	&94.27 	&81.85 	&73.64 	&88.69 	&92.25 	&78.32 	&69.63 	&85.90 	&89.46   \\
w/o $\mathcal{L}_C$+$\mathcal{M}_C$	&91.63 	&87.54 	&\uline{95.12} 	&96.70 	&87.00 	&81.08 	&91.89 	&94.33 	&78.32 	&69.63 	&85.90 	&89.46   \\
\hline
w/ InfoNCE~\citep{Kaiming_2020}	&90.54 	&86.28 	&94.06 	&96.03 	&85.29 	&79.06 	&90.26 	&93.07 	&79.87 	&72.62 	&85.61 	&88.74  \\
w/ MCLET~\citep{Zhiwei_2023}	&91.85 	&87.79 	&\textbf{95.20} 	&\textbf{96.77} 	&87.69 	&81.78 	&\uline{92.67} 	&\uline{95.03} 	&79.95 	&72.18 	&86.00 	&89.61  \\
\hline
${\rm M^3EL}$   &\textbf{92.30} 	&\textbf{88.84} 	&\textbf{95.20} 	&\uline{96.71} 	&\uline{88.26} 	&82.82 	&\textbf{92.73} 	&\textbf{95.34} 	&\textbf{81.29} 	&\textbf{74.06} 	&\uline{86.57} 	&\textbf{90.04} \\
\bottomrule
\end{tabular*}
\label{table_ablation_studies}
\end{table*}

\smallskip
\noindent\textbf{$\triangleright$ Impact of Different Losses.}
\vic{The upper part of Table~\ref{table_ablation_studies} shows the individual contribution of different losses. We can observe that remove any loss will generally bring certain degree of performance degradation. On the RichpediaMEL and WikiDiverse datasets, the performance loss caused by removing the visual loss is higher than by removing the textual loss. However, on the WikiMEL dataset, the situation is the opposite, i.e., the performance fluctuation caused by removing the textual loss is larger. This shows that the richness of modal knowledge is the main factor that determines the impact of the corresponding modality on model performance. In addition, removing the contrastive loss  also has certain impact on performance, especially on the WikiDiverse dataset. 
This is mainly due to the fact that we also consider the negative examples within a modality  in the contrastive learning process to obtain better discriminative modal embedding representations. Although there are some singularies in some indicators of some datasets when $\mathcal{L}_T$, $\mathcal{L}_V$, $\mathcal{L}_C$ are removed, this is  acceptable because the information richness of textual and visual modalities in each dataset is different. Indiscriminately removing modules for specific modalities will cause unknown performance losses. The overall performance is the best when all losses and modules are simultaneously used.}


\smallskip
\noindent\textbf{$\triangleright$ Impact of Different Modules.} 
\vic{The middle part of Table~\ref{table_ablation_studies} shows the individual contribution of different modules. It should be mentioned that removing the corresponding module involves removing two aspects: for example, when  removing the cross-modal matching network (CMN),  the matching score $\mathcal{M}_C$ related to the CMN module needs to be removed from the union matching score $\mathcal{M}_U$, and  the loss $\mathcal{L}_C$ introduced by $\mathcal{M}_C$ also needs to be removed. Since  the  loss is calculated from the matching score of the corresponding module, there will be no corresponding loss value without a matching score. From the experimental results in Table~\ref{table_ablation_studies}, we can observe that when compared to only removing the loss value $\mathcal{L}_C$ related to the corresponding module, removing the matching score of the that  module from the union matching score will  lead to a greater performance degradation. Taking the textual modality of the WikiMEL dataset as an example, if only the textual modality loss is removed, the corresponding MRR and Hits@1 metrics  respectively are 91.64 and 87.60. After further removing the matching score $\mathcal{M}_T$ related to the textual modality, the corresponding MRR and Hits@1 metrics become 90.22 and 86.40, falling by 1.42\% and 1.2\% respectively. A similar situation occurrs when modules related to visual modalities and cross-modalities are removed, which  confirms the usefulness of  each module.}


\smallskip
\noindent\textbf{$\triangleright$ Impact of Various Contrastive Loss.} \vic{The lower part of Table~\ref{table_ablation_studies} shows the experimental results of replacing the ICL module with the  InfoNCE~\citep{Kaiming_2020} and MCLET~\citep{Zhiwei_2023}  contrastive losses. We can observe that with our ICL's ${\rm M^3EL}$ can achieve better experimental results in most cases. Specifically, taking  the WikiMEL dataset as an example, compared with InfoNCE and MCLET, the ICL module improves the MRR metric by 1.84\% and 0.45\%, respectively. This is due to the fact that we consider both inner-source and inter-source  negative samples in the ICL module. In addition, we introduce a weight coefficient in Equation~\ref{equation_2} to reconcile the possible imbalance of negative samples. We conduct parameter analysis experiments on the effect of inner-source and inter-source alignment weights $\beta$ and $\gamma$ in \textcolor{blue}{\textbf{Appendix \hyperlink{parameter_sensitivity}{A}}}.}

\begin{table}[!htp]
\setlength{\abovecaptionskip}{0.18cm}
\renewcommand\arraystretch{1.3}
\setlength{\tabcolsep}{0.36em}
\centering
\small
\caption{Evaluation of ablation studies of different pooling operation on WikiDiverse dataset.}
\begin{tabular*}{0.67\linewidth}{@{}ccccc@{}}
\bottomrule
\multicolumn{1}{c}{\multirow{2}{*}{\textbf{Methods}}} & \multicolumn{4}{c}{\textbf{WikiDiverse }} \\
\cline{2-5}
& \textbf{MRR} & \textbf{Hits@1} & \textbf{Hits@3} & \textbf{Hits@5} \\
\hline
\textit{max}   &77.28  &68.86  &83.73  &87.63 \\ 
\textit{soft}   &77.39  &69.01  &83.78  &87.82 \\ 
\textit{mean}  &\textbf{81.29}  &\textbf{74.06}  &\textbf{86.57}  &\textbf{90.04} \\ 
\hline
\end{tabular*}
\label{table_attention_operation}
\end{table}

\smallskip
\noindent\textbf{$\triangleright$ Effect of Different Pooling Operation in Equation~\ref{equation_5}.} The mean pooling operation in Equation~\ref{equation_5} is mainly used to reduce the spatial dimension of $\alpha^L$ to facilitate its calculation with $\textbf{T}_e^G$ to obtain the global-to-local matching score $\mathcal{M}_T^{G2L}$. We replace the mean pooling operation with max pooling and soft pooling~\citep{Weiran_2021} operations on the WikiDiverse dataset to  study the impact of the pooling operation. The corresponding results are shown in  Table~\ref{table_attention_operation}. We find that the advantage of using mean pooling operation is more prominent. A possible explanation is that it is more effective to comprehensively consider the representation of each token in a sentence or each patch in an image than to consider only some important tokens or patches.

\begin{table}[!htp]
\setlength{\abovecaptionskip}{0.03cm}
\renewcommand\arraystretch{1.2}
\setlength{\tabcolsep}{0.3em}
\centering
\small
\caption{Amount of parameters and calculation with various contrastive loss.}
\begin{tabular*}{0.9\linewidth}{@{}ccccc@{}}
\bottomrule
\multicolumn{1}{c}{\textbf{Metrics}}&\multicolumn{1}{c}{\textbf{Mode}}&\multicolumn{1}{c}{\textbf{WikiMEL}}&\multicolumn{1}{c}{\textbf{RichpediaMEL}}&\multicolumn{1}{c}{\textbf{WikiDiverse}} \\
\hline
\multirow{2}{*}{\#FLOPs} & InfoNCE &1.469G &1.469G &2.233G \\
& MCLET &1.402G &1.402G &2.128G \\
& ICL &\textbf{1.369G} &\textbf{1.369G} &\textbf{2.076G} \\
\hline
\multirow{2}{*}{\#Params} & InfoNCE &335.259K &335.259K &324.753K \\
& MCLET &321.423K &321.423K &311.721K \\
& ICL &\textbf{314.529K} &\textbf{314.529K} &\textbf{305.217K} \\
\bottomrule
\end{tabular*}
\label{table_flops_and_params}
\end{table}

\smallskip
\noindent\textbf{$\triangleright$ Resource Consumption Using Various Contrastive Loss.} We analyze the resource consumption from two perspectives: time complexity and space complexity. The corresponding results on three different datasets are shown in the Table~\ref{table_flops_and_params}. The time complexity is measured by the number of floating-point operations (\#FLOPs) required during training phase, while the space complexity refers to the amount of a models parameters (\#Params). The larger the values of \#FLOPs and \#Params are, the more computing power and higher memory usage are required during the training process. We observe that compared with InfoNCE and MCLET, ICL requires less time and space overhead in each dataset. Specifically, on the WikiMEL dataset, ICL reduces \#FLOPs and \#Params by 2.4\% and 2.1\%, respectively, compared to MCLET, which demonstrates that ICL is more efficient.


\section{Conclusion}

In this paper, we propose ${\rm M^3EL}$, a multi-level matching network for multimodal entity linking. ${\rm M^3EL}$ simultaneously considers the diversity of negative samples from the same modality and bidirectional cross-modality interaction. Specifically, we introduce an  intra-modal contrastive loss to obtain better discriminative representations that are faithful to certain modality. Furthermore, we design  intra-modal and inter-modal matching mechanisms to explore multi-level multimodal interactions. Extensive experiments on three datasets demonstrate ${\rm M^3EL}$'s robust performance.

\section*{Acknowledgments}
This work has been supported by the National Natural Science Foundation of China (No.61936012, No.62076155), by the Key Research and Development Project of Shanxi Province (No.202102020101008), by the Science and Technology Cooperation and Exchange Special Project of Shanxi Province (No.202204041101016).

\bibliographystyle{ACM-Reference-Format}
\bibliography{sample-base}

\section*{Appendix}

\subsection*{A\,\,\,Parameter Sensitivity}
\begin{figure*}[htbp]
  \centering\subcaptionbox{Numbers of Heads $K$}{
  \includegraphics[width=0.238\linewidth]{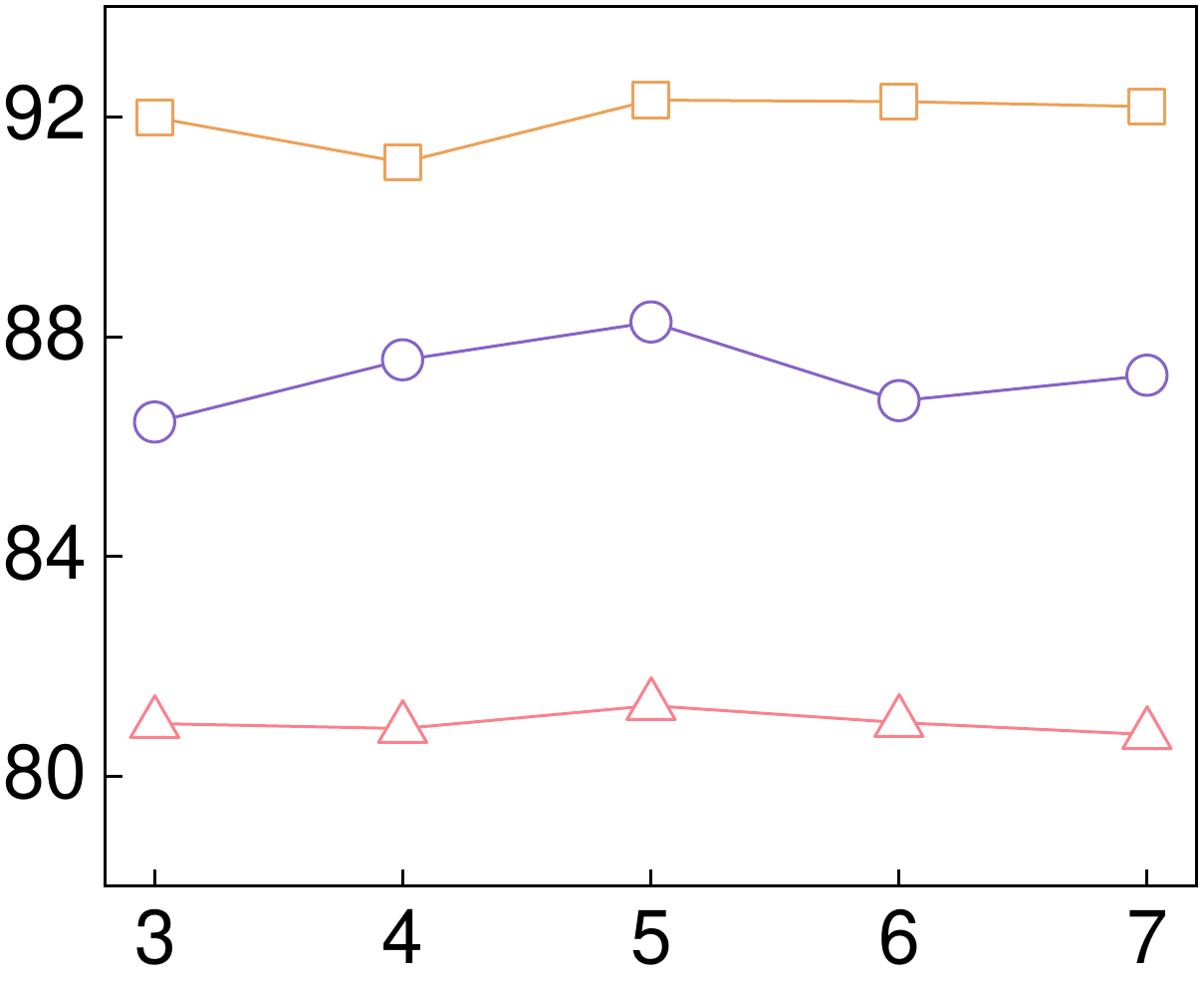}}\hfill
  \subcaptionbox{Temperature $\tau$}{
  \includegraphics[width=0.242\linewidth]{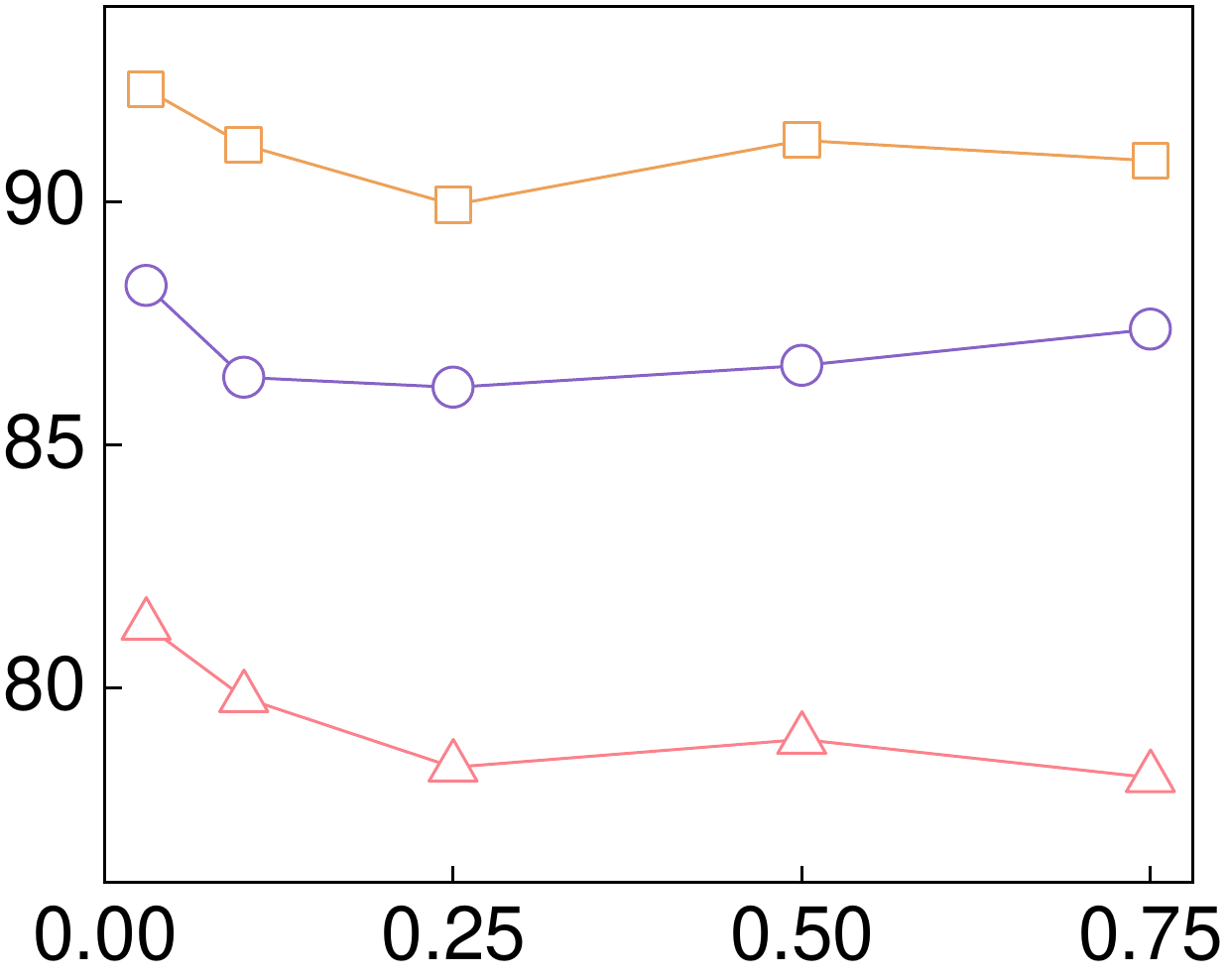}}\hfill
  \subcaptionbox{Inner-souce Weight $\beta$}{
  \includegraphics[width=0.24\linewidth]{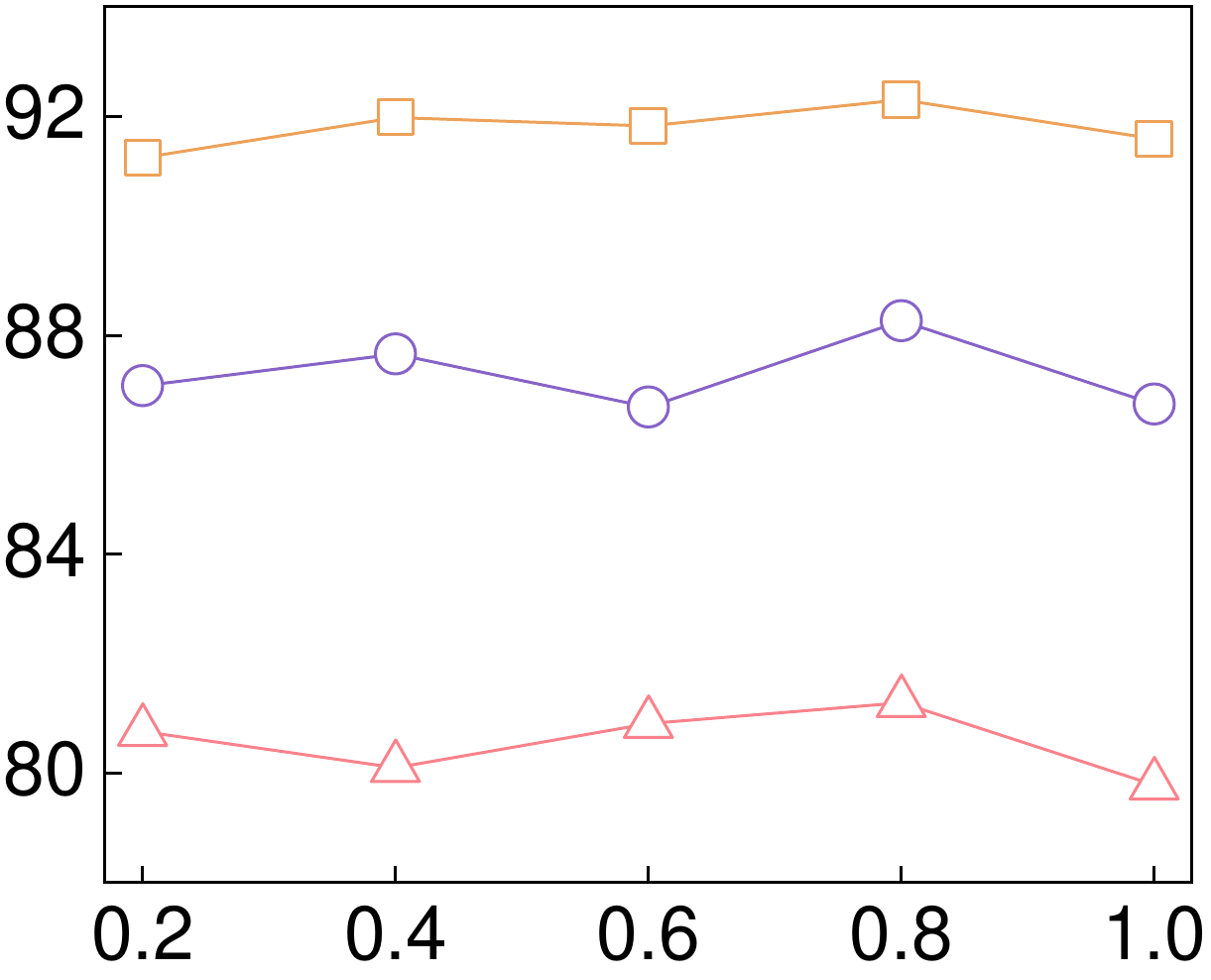}}\hfill
  \subcaptionbox{Inter-souce Weight $\gamma$}{
  \includegraphics[width=0.24\linewidth]{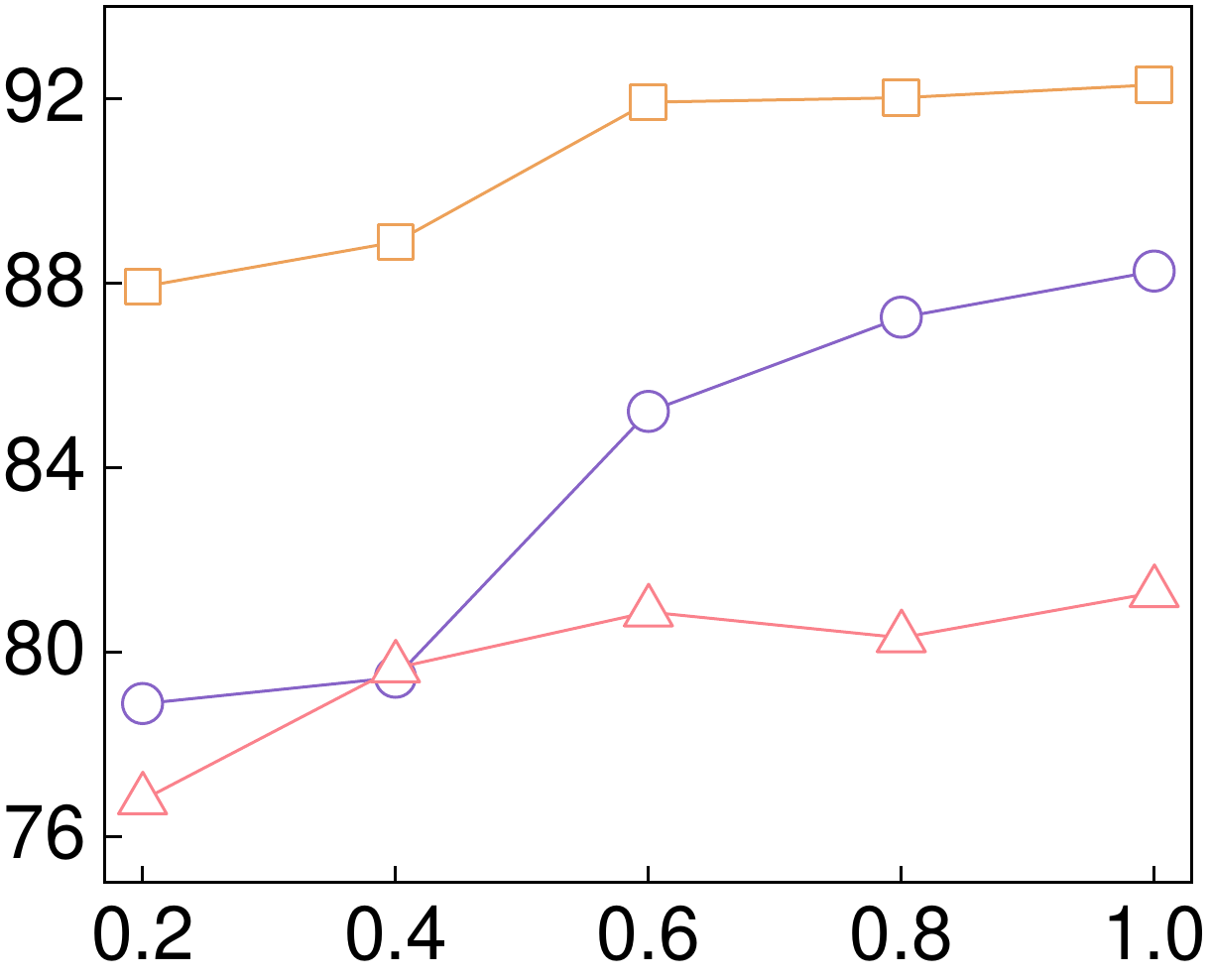}}

  \includegraphics[width=0.44\linewidth]{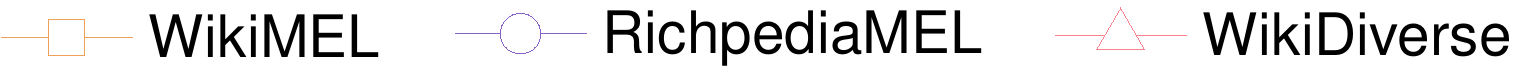}
  \caption{Parameter sensitivity experiments under different conditions on WikiMEL, RichpediaMEL and WikiDiverse datasets.}
\label{figure_hyper_parameters}
\end{figure*}

\hypertarget{parameter_sensitivity}{}
\vic{We carry out parameter sensitivity experiments on the WikiMEL, RichpediaMEL and WikiDiverse datasets. The results are shown in Figure~\ref{figure_hyper_parameters}, including: \textit{a}) effect of numbers of heads $K$; \textit{b}) effect of temperature coefficient $\tau$; \textit{c}) effect of inner-source and inter-source alignment weight $\beta$ and $\gamma$.}

\smallskip
\noindent\textbf{$\triangleright$ Effect of Numbers of Heads $K$.} \vic{In Equation~\eqref{equation_7}, we use a multi-head attention mechanism to achieve the fusion between cross-modal local features and the global feature. In Figure~\ref{figure_hyper_parameters}(a), we observe that the number of multi-head attention heads  has a greater impact on the WikiMEL and RichpediaMEL datasets than on  WikiDiverse. We achieve the best performance  on all three datasets when the number is set to 5.}


\smallskip
\noindent\textbf{$\triangleright$ Effect of Temperature Coefficient $\tau$.} 
\vic{The temperature coefficient can be used to adjust the similarity measure between samples. When the temperature coefficient is higher, the model is more likely to reduce the difference between positive samples and negative samples. This  can cause the learned feature representation to be less sensitive to the differences between different samples. When the temperature is lower, the model pays more attention to subtle differences between samples. This  may lead to being overly sensitive to noise and local changes, thereby reducing the model's generalization ability. Therefore, selecting an appropriate temperature coefficient value requires appropriate adjustments based on the characteristics of a dataset. We can see in Figure~\ref{figure_hyper_parameters}(b) that selecting a smaller temperature value is beneficial to all three datasets and the best results are achieved when the temperature value is set to 0.03.}


\smallskip
\noindent\textbf{$\triangleright$ Effect of Inner-source and Inter-source Alignment Weights $\beta$ and $\gamma$.} 
\vic{The weight coefficients $\beta$ and $\gamma$ in Equation~\eqref{equation_2} can be used to control the importance of negative samples from inner-source and inter-source. An appropriate adjustment of the weights values, based on the dataset, can help  avoiding the performance impact caused by improper selection of negative samples. We can observe in Figure~\ref{figure_hyper_parameters}(c) and (d) that selecting a larger $\gamma$ value is more beneficial to the three datasets. However, there is no rule to follow for the selection of the $\beta$ value because as $\beta$ increases, the performance of the three datasets fluctuates.}

\begin{table*}[!htp]
\setlength{\abovecaptionskip}{0.18cm}
\renewcommand\arraystretch{1.3}
\setlength{\tabcolsep}{0.36em}
\centering
\small
\caption{Evaluation of ablation studies of T2V and V2T mechanism on WikiMEL, RichpediaMEL and WikiDiverse datasets.}
\begin{tabular*}{0.85\linewidth}{@{}ccccccccccccc@{}}
\bottomrule
\multicolumn{1}{c}{\multirow{2}{*}{\textbf{Methods}}} & \multicolumn{4}{c}{\textbf{WikiMEL}} & \multicolumn{4}{c}{\textbf{RichpediaMEL}} & \multicolumn{4}{c}{\textbf{WikiDiverse}}\\
\cline{2-5}\cline{6-9}\cline{10-13}
& \textbf{MRR} & \textbf{Hits@1} & \textbf{Hits@3} & \textbf{Hits@5} & \textbf{MRR} & \textbf{Hits@1} & \textbf{Hits@3} & \textbf{Hits@5} & \textbf{MRR}  & \textbf{Hits@1} & \textbf{Hits@3} & \textbf{Hits@5} \\
\hline
\textit{w/} T2V   &91.53  &88.08  &94.16  &95.65   &86.53 &81.11  &90.82  &93.40  &76.08  &68.05  &82.00  &86.14 \\ 
\textit{w/} V2T   &91.80  &88.24  &94.76  &96.32   &87.89 &81.96  &91.75  &94.27  &76.73  &67.95  &84.12  &87.58 \\ 
\textit{w/} T2V \& V2T  &\textbf{92.30}  &\textbf{88.84}  &\textbf{95.20}  &\textbf{96.71}   &\textbf{88.26} &\textbf{82.82}  &\textbf{92.73}  &\textbf{95.34}  &\textbf{81.29}  &\textbf{74.06}  &\textbf{86.57}  &\textbf{90.04} \\ 
\hline
\end{tabular*}
\label{table_effective_bidirection_mechanism}
\end{table*}

\begin{table*}[!htp]
\setlength{\abovecaptionskip}{0.18cm}
\renewcommand\arraystretch{1.3}
\setlength{\tabcolsep}{0.36em}
\centering
\small
\caption{Evaluation of ablation studies of different feature extractor on WikiMEL, RichpediaMEL and WikiDiverse datasets.}
\begin{tabular*}{0.82\linewidth}{@{}ccccccccccccc@{}}
\bottomrule
\multicolumn{1}{c}{\multirow{2}{*}{\textbf{Methods}}} & \multicolumn{4}{c}{\textbf{WikiMEL}} & \multicolumn{4}{c}{\textbf{RichpediaMEL}} & \multicolumn{4}{c}{\textbf{WikiDiverse}}\\
\cline{2-5}\cline{6-9}\cline{10-13}
& \textbf{MRR} & \textbf{Hits@1} & \textbf{Hits@3} & \textbf{Hits@5} & \textbf{MRR} & \textbf{Hits@1} & \textbf{Hits@3} & \textbf{Hits@5} & \textbf{MRR}  & \textbf{Hits@1} & \textbf{Hits@3} & \textbf{Hits@5} \\
\hline
\textit{BLIP}   &58.05  &51.13  &62.00  &65.76   &41.55 &33.13  &45.90  &50.95  &36.85  &27.19  &41.92  &49.23 \\
\textit{CLIP}  &\textbf{92.30}  &\textbf{88.84}  &\textbf{95.20}  &\textbf{96.71}   &\textbf{88.26} &\textbf{82.82}  &\textbf{92.73}  &\textbf{95.34}  &\textbf{81.29}  &\textbf{74.06}  &\textbf{86.57}  &\textbf{90.04} \\ 
\hline
\end{tabular*}
\label{table_effective_feature_extractor}
\end{table*}

\begin{table*}[!htp]
\setlength{\abovecaptionskip}{0.18cm}
\renewcommand\arraystretch{1.3}
\setlength{\tabcolsep}{0.36em}
\centering
\small
\caption{Evaluation of ablation studies of independent mechanism in IMN module on WikiMEL, RichpediaMEL and WikiDiverse datasets.}
\begin{tabular*}{0.83\linewidth}{@{}ccccccccccccc@{}}
\bottomrule
\multicolumn{1}{c}{\multirow{2}{*}{\textbf{Methods}}} & \multicolumn{4}{c}{\textbf{WikiMEL}} & \multicolumn{4}{c}{\textbf{RichpediaMEL}} & \multicolumn{4}{c}{\textbf{WikiDiverse}}\\
\cline{2-5}\cline{6-9}\cline{10-13}
& \textbf{MRR} & \textbf{Hits@1} & \textbf{Hits@3} & \textbf{Hits@5} & \textbf{MRR} & \textbf{Hits@1} & \textbf{Hits@3} & \textbf{Hits@5} & \textbf{MRR}  & \textbf{Hits@1} & \textbf{Hits@3} & \textbf{Hits@5} \\
\hline
\textit{independent}   &91.90  &88.33  &94.64 &96.34   &87.53 &82.34  &91.61  &94.08  &79.13  &71.27  &85.23  &89.12 \\
\textit{same}  &\textbf{92.30}  &\textbf{88.84}  &\textbf{95.20}  &\textbf{96.71}   &\textbf{88.26} &\textbf{82.82}  &\textbf{92.73}  &\textbf{95.34}  &\textbf{81.29}  &\textbf{74.06}  &\textbf{86.57}  &\textbf{90.04} \\ 
\hline
\end{tabular*}
\label{table_effective_same_mechanism}
\end{table*}

\subsection*{B\,\,\,Additional Experiments}
\hypertarget{additional_experiments}{}

\smallskip
\noindent\textbf{$\triangleright$ Effect of Bidirectional Interaction Mechanism.} We conduct further ablation experiments on the T2V and V2T matching mechanisms of the CMN module in Section ~\ref{section_cmn} on three datasets. The corresponding results are shown in  Table~\ref{table_effective_bidirection_mechanism}. We can observe that using only the T2V or V2T matching mechanism will bring a certain degree of performance degradation on the three datasets. Specifically, the performance loss caused by using only T2V is greater than that of using only V2T. Furthermore, on the WikiDiverse dataset, the performance degradation is the largest after removing T2V or V2T, which fully demonstrates the necessity of the two mechanisms. We further give the rational behind the existence of the two mechanisms. First, we need to give some explanations of terms. As described in Section~\ref{section_multimodal_embedding}, global textual feature refer to the embedding representation of the entire textual modality sentence, local textual features refers to the embedding representation of each token in the textual modality sentence, global visual feature refers to the embedding representation of the entire visual modality image, and local visual features refer to the embedding representation of each patch in the visual modality image. Secondly, for T2V, its input is global-textual feature and local visual features, \textit{i.e.,} using the global representation at the sentence level to supervise the local representation at the image patch level, which is suitable for scenes with richer textual  semantic knowledge, while V2T's input is global visual feature and local textual features, which is more suitable for scenes with more  visual information from the image. The T2V and V2T mechanisms are applicable to different scenarios. The combination of the two can complement each other and be more robust to changes in scenarios.

\smallskip
\noindent\textbf{$\triangleright$ Effect of Different Feature Extractor.} We use CLIP as the feature extractor in MFE, mainly considering the following two factors: On the one hand, CLIP has good performance. On the other hand, the memory size of our experimental machine is limited, making it difficult to run larger models. In principle, CLIP can be replaced by any encoder that can obtain textual and visual modality embeddings, such as BLIP~\citep{Junnan_2022} or BLIP-2~\citep{Junnan_2023}. Considering the current status of our experimental machine, we choose BLIP-vqa-base to replace CLIP for the experiment. The corresponding results are shown in Table~\ref{table_effective_feature_extractor}. We find that using BLIP actually obtains worse results, the main reason is that in order to run the experiment, we performed float16 quantization on BLIP-vqa-base (otherwise the memory will overflow), which will lead to a large degree of precision loss. We leave experiments with full-scale runs of BLIP and BLIP-2 as future work.

\smallskip
\noindent\textbf{$\triangleright$ Effect of  Independent Mechanism in IMN Module.} To verify the performance of the independent mechanism proposed by the MIMIC~\citep{Pengfei_2023} model on our ${\rm M^3EL}$ model, we conduct experiments on the WikiMEL, RichpediaMEL, and WikiDiverse datasets. The corresponding results are shown in Table~\ref{table_effective_same_mechanism}. We observe that after replacing our unified mechanism with an independent mechanism, there is a certain degree of performance loss on the three datasets, which illustrates that the independent mechanism is not suitable for ${\rm M^3EL}$. A possible explanation is that the interaction mechanism within the modality is similar, and there is no need to customize a unique interaction system for each modality. Instead, a simpler approach is more appropriate.

\end{document}